\documentclass{article}

\PassOptionsToPackage{numbers}{natbib}
\usepackage[main,final]{neurips_2026}

\usepackage[utf8]{inputenc}
\usepackage[T1]{fontenc}
\usepackage{hyperref}
\usepackage{url}
\usepackage{booktabs}
\usepackage{tabularx}
\usepackage{graphicx}
\usepackage{array}
\usepackage{amsmath}
\usepackage{amsfonts}
\usepackage{algorithm}
\usepackage{algpseudocode}
\usepackage{nicefrac}
\usepackage{microtype}
\usepackage[table]{xcolor} %
\usepackage{xcolor}
\usepackage{makecell}   %
\usepackage{tcolorbox}
\usepackage{multirow}
\usepackage{multicol}
\usepackage{wrapfig}
\usepackage{caption}
\tcbuselibrary{skins,breakable}
\usepackage{enumitem}

\definecolor{algblue}{HTML}{1F5E8C}
\definecolor{algorange}{HTML}{C16A13}
\definecolor{alggreen}{HTML}{3C7A3E}
\definecolor{algline}{HTML}{BFBFBF}

\usepackage{kotex}

\definecolor{tblBest}{HTML}{DDEBDD}      %
\definecolor{tblSecond}{HTML}{E3ECF6}    %
\definecolor{tblOurs}{HTML}{F7E6D0}      %
\definecolor{tblHeader}{HTML}{F2F2F2}    %

\newcommand{\best}[1]{\cellcolor{tblBest}{{\bfseries\boldmath #1}}}
\newcommand{\second}[1]{\cellcolor{tblSecond}{\underline{#1}}}

\newcommand{\score}[2]{$#1$ {\scriptsize $\pm\,#2$}}

\newcommand{\STAB}[1]{\begin{tabular}{@{}c@{}}#1\end{tabular}}

\title{Multi-Agent Coordination\\via Support-Preserving Distillation}

\author{%
  Sangmin Lee \quad Youngju Na \quad Chanmi Lee \quad Sung-eui Yoon\textsuperscript{$\dagger$} \\
  School of Computing, KAIST \\
  \texttt{\{alex6095,yjna2907,chanmi99,sungeui\}@kaist.ac.kr} \\
  {\normalfont\footnotesize \textsuperscript{$\dagger$}Corresponding author} \\[4pt]
  {\normalfont\small\hypersetup{hidelinks}%
   \href{https://alex6095.github.io/mosdot/}{\textcolor{algblue}{\nolinkurl{https://alex6095.github.io/mosdot/}}}}
}

\begin{document}

\maketitle

\begin{abstract}
Offline MARL increasingly relies on
generative policies to model multimodal joint behavior, typically by
distilling a centralized teacher into decentralized one-step actors
under the CTDE. We identify a failure mode at the teacher training stage:
standard flow-based teachers pair noise with replay targets
independently, so nearby noise samples can be routed toward conflicting
coordination modes. The teacher then produces samples between valid
modes, and because the distillation loss regresses each local actor
onto the conditional mean of the teacher's output given local input,
this error is not absorbed but propagated to the student.
To remove this teacher-side artifact, we propose
\emph{Mode-Support Semi-Discrete Optimal Transport} (MoSDOT), which
summarizes multimodal replay into a finite mode support with prescribed
capacities and uses conditional semi-discrete optimal transport to
assign each noise sample to a single mode before teacher training. We
additionally study a shared-randomness variant that uses a shared noise
component at execution to expose the residual gap intrinsic to
strict-product execution. On controlled diagnostics and offline MARL
benchmarks, MoSDOT improves endpoint quality and routing consistency,
particularly on datasets exhibiting multimodal joint behavior.
\end{abstract}

\section{Introduction}

Offline multi-agent reinforcement learning (MARL) aims to learn coordinated
policies from a fixed dataset, without further environment interaction. The
core difficulty is that coordination is defined over the \emph{joint action}
distribution of all agents: individually plausible per-agent actions may
combine into unseen joint actions that violate safety or yield low value.
This is especially problematic when the data contains multiple valid
coordination modes, defined as distinct joint behavior patterns, where a
unimodal policy averages alternatives into an invalid in-between action.
Expressive generative policies based on diffusion and flow matching address
this multimodality by modeling joint action as a multi-mode
distribution~\cite{li2025dof,lee2026macflow,pang2026vgm2p}, but their
multi-step sampling can be costly at decision time and complicate
value-guided actor extraction through critic gradients~\cite{park2025fql}.
A recent line of work therefore trains a centralized generative
\emph{teacher} over joint actions and \emph{distills} it into decentralized
one-step \emph{students}, which act locally and admit more direct
value-guided extraction~\cite{lee2026macflow}.

\begin{figure}[t]
  \centering
  \includegraphics[width=0.95\linewidth]{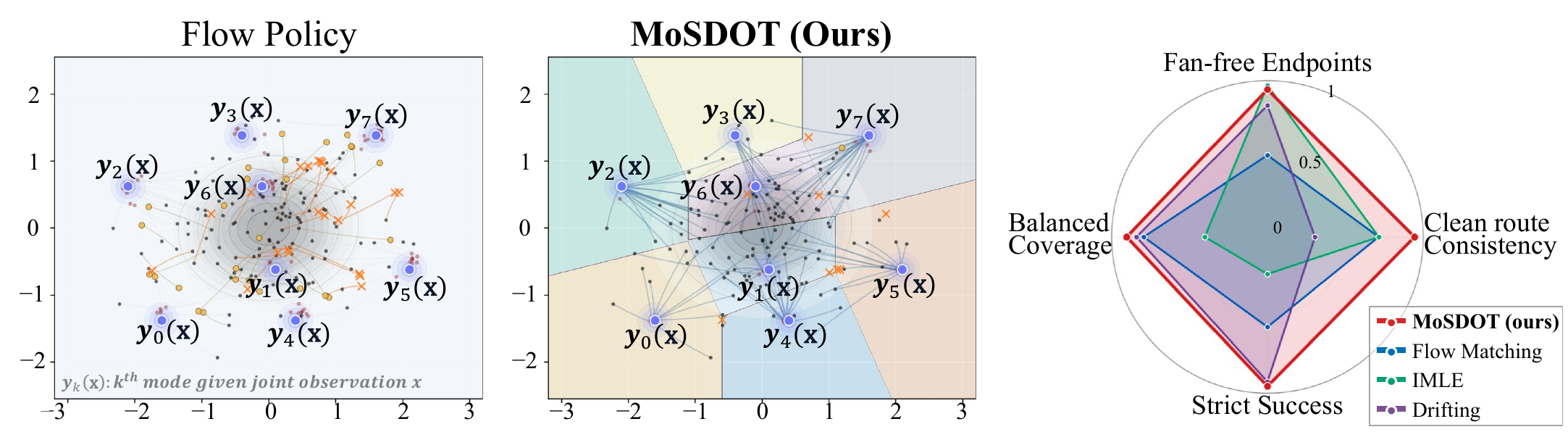}
  \vspace{-6pt}
  \caption{
  \textbf{Overview of source-to-mode routing.}
  \textbf{(Left) Standard Flow Policy:} independent noise--target
  pairing can entangle nearby source samples with incompatible
  coordination modes, producing off-support fan mass.
  \textbf{(Middle) MoSDOT:} semi-discrete optimal transport partitions
  noise space into capacity-matched cells assigned to coordinated
  behavior modes.
  \textbf{(Right)} Route quality is measured by off-support avoidance
  (\textit{fan-free endpoints}), consistency of source-to-mode routing (\textit{clean route consistency}),
  endpoint accuracy to valid modes (\textit{strict success}), and balanced mass
  across modes (\textit{balanced coverage}). Full metric definitions are provided in
  Appendix~\ref{app:metric_defs}.
}
  \label{fig:joint_support_factorization_failure}
  \vspace{-\baselineskip}
\end{figure}

Distillation, however, exposes a gap between joint modeling and decentralized
execution. The teacher conditions on the full joint context and a shared
noise sample, whereas each local actor sees only its own context and
independent noise. If the teacher's choice between coordination modes depends
on information hidden from a local actor, the $\ell_2$ distillation loss
regresses to the conditional mean and collapses distinct modes into an
invalid in-between action. Mode-preserving distillation therefore requires
not only that the teacher cover the target modes, but also that those modes
be \emph{compatible with the information available to local actors}. This
makes coherent teacher training critical: standard flow-based training
pairs noise samples with replay targets independently, leaving an
unstructured mapping between noise and joint behaviors. With multimodal
targets, adjacent noise points are forced to learn conflicting modes
simultaneously, producing trajectories that pass through low-density regions
or form spurious bridges between modes
(Fig.~\ref{fig:joint_support_factorization_failure}, left). Because
independent local policies capture only the conditional mean, even minor
mode-mixing in the teacher can lead the student to collapse onto invalid
joint actions. A clean teacher alone, however, is not sufficient. Independent agents cannot
in general represent correlated joint support such as the anti-correlated
XOR target, a residual mismatch we call the \emph{product-projection gap}.

Our \emph{Mode-Support Semi-Discrete Optimal Transport} (MoSDOT) addresses
the teacher-side problem by summarizing multimodal replay into a finite set
of representative joint behavior modes — the \emph{mode support} — with
prescribed capacities, and partitioning the noise space via semi-discrete
optimal transport into cells assigned to those modes before teacher training
(Fig.~\ref{fig:joint_support_factorization_failure}, middle). 
MoSDOT thus reduces teacher-side routing artifacts and preserves the mode support that local actors can
recover, while leaving the product-projection gap explicit. To investigate
this residual gap, we additionally consider a shared-randomness variant in
which a portion of the noise input acts as an execution-time shared signal:
all agents read the same coordination mode from this signal but still
compute actions from their own observations and local noise, allowing
correlated joint behavior without exchanging observations or actions.

In controlled diagnostics, MoSDOT mitigates teacher-side routing
artifacts and improves source-to-mode assignment consistency.
Fig.~\ref{fig:joint_support_factorization_failure} (right) summarizes
these gains across endpoint quality, route consistency, strict success,
and coverage balance. On offline MARL benchmarks, MoSDOT performs
competitively with existing generative baselines, particularly on
datasets containing diverse joint behaviors. Overall, MoSDOT separates
teacher-side source-routing artifacts from the residual
product-projection gap, and provides a measurable interface between the
centralized training and decentralized execution (CTDE)~\cite{wang2023offline}.

\section{Related Work}

\vspace{-4pt}
\subsection{Generative Policies and Distillation in Offline MARL}
Offline reinforcement learning (RL) aims to learn policies from fixed datasets
by constraining policy improvement close to actions represented in the data
to avoid value overestimation~\cite{fujimoto2019off,kumar2020conservative,
fujimoto2021minimalist,kostrikov2021offline}. In multi-agent settings, the
support is defined over coordinated \emph{joint} actions, so small per-agent
deviations can combine into joint behaviors absent from replay. Offline
MARL methods address this through implicit
constraints~\cite{yang2021believe}, actor
rectification~\cite{pan2022plan}, and global-to-local value regularization
under the CTDE. Standard actor or value-factorized
parameterizations rarely model multimodal joint action support explicitly,
so distinct coordination modes can be averaged into incoherent in-between
actions.

Expressive generative policies address this multimodality. Diffusion and
flow-based models have been used for trajectory-level
planning~\cite{janner2022planning,ajay2022conditional}, action
generation~\cite{wang2022diffusion}, and one-step offline policy
learning~\cite{park2025fql}. In offline MARL,
MADiff~\cite{zhu2024madiff} models coordinated multi-agent trajectories
with diffusion, and DoF~\cite{li2025dof} factorizes diffusion generation
and formulates Individual-Global-identically-Distributed (IGD) generation
as a distributional analogue of value factorization. These approaches
improve expressiveness and execution efficiency, but the quality of the
centralized generative distribution remains an important prerequisite.

Policy distillation transforms multimodal teachers into efficient one-step
students. Since pointwise regression tends to average modes, recent work
has introduced set-level objectives based on Implicit Maximum Likelihood
Estimation (IMLE) and Chamfer matching~\cite{dong2026flowtoonestep}, as
well as drifting objectives that shape support through attraction toward
data and repulsion among generated samples~\cite{deng2026drifting}. In
offline RL, FQL~\cite{park2025fql} distills a behavioral flow into a
Q-regularized one-step policy, and MAC-Flow~\cite{lee2026macflow}
distills a centralized multi-agent flow teacher into decentralized
one-step actors. These methods primarily focus on the distillation process
itself, while our work instead studies the upstream construction of the
teacher distribution before distillation begins.

\vspace{-4pt}
\subsection{Mode Alignment in Flow Policies and Factorized Execution}
\vspace{-4pt}
Flow-based generative models are often sensitive to how source noise is
coupled to data during training. Flow matching trains continuous
normalizing flows by regressing vector fields along conditional probability
paths~\cite{lipman2022flow}. With independent noise--target pairings,
multimodal targets can produce samples that fall between valid modes.
Optimal transport (OT) flow matching~\cite{tong2023improving} mitigates this effect by coupling noise and data
through transport plans, and AlignFlow~\cite{kong2025alignflowimproving} uses
semi-discrete optimal transport (SDOT) to partition source space into
Laguerre cells assigned to empirical data samples for image
generation.

A related issue arises in MARL, where centralized training must be
compatible with decentralized execution. Value factorization methods
such as VDN~\cite{sunehag2017value}, QMIX~\cite{rashid2020monotonic},
and QTRAN~\cite{son2019qtran} impose structure on centralized value
functions so that decentralized action selection remains aligned with
a global objective. Generative offline MARL raises a distributional
analogue of this question. A centralized teacher may represent
multimodal joint behavior, but the distilled local actors see only
local information and are often trained with pointwise regression
losses.

Our work focuses on the corresponding upstream failure mode. Before
student training, independent teacher pairings can create off-support
or entangled source routes between coordinated joint modes. During
distillation, mode variation that is not locally observable is further
averaged by the student objective. MoSDOT reduces this teacher-side
routing artifact by aligning source-noise regions with the mode
support via conditional SDOT, providing cleaner teacher targets for
support-preserving decentralized distillation.

\section{Method}
\label{sec:method}
\subsection{Problem Setup}
\label{subsec:problem_setup}

We consider cooperative offline MARL under a decentralized partially observable
Markov decision process (Dec-POMDP)~\cite{bernstein2002complexity}, defined as
$\mathcal M = \langle \mathcal{N}, \mathcal S, \mathcal O, \mathcal{A}, P, Z, R, \gamma \rangle$.
Let $\mathcal{N} = \{1,...,N\}$ be the set of $N$ agents, with $\mathcal{S}$, $\mathcal{O}=\{\mathcal O_i\}_{i=1}^N$, and $\mathcal{A}=\{\mathcal A_i\}_{i=1}^N$ denoting the joint global state, observation, and action spaces, respectively.
We denote $\mathbf{x}=(o_1,\ldots,o_N) \in \mathcal{O} $ as the joint observation context and $\mathbf a=(a_1,\ldots,a_N) \in \mathcal{A}$ as the joint action.
$P(s'|s,\mathbf{a}):\mathcal{S}\times\mathcal{A}\rightarrow\Delta(\mathcal{S})$ is the 
state transition function and $Z^i(o_i|s):\mathcal{S}\rightarrow\Delta(\mathcal{O}_i)$ 
is the $i$-th agent's observation function, where $s\in \mathcal{S}$ and $\Delta(\cdot)$ denotes the set of probability measures over the given space.
The team reward is generated based on the reward function $R(s,\mathbf a, s') : \mathcal{S}\times\mathcal{A}\times\mathcal{S} \rightarrow \mathbb{R}$, and $\gamma$ is a discount factor.

At each timestep, all decentralized agents choose their actions $\mathbf{a}$ conditioned on their local observations $\mathbf{x}$ provided by the environment in state $s$.
The environment then transitions to $s'\sim P(s'|s,\mathbf{a})$, and the team reward is given by $R(s,\mathbf a, s')$.
In the offline setting, learning uses a fixed dataset $\mathcal D$ of joint transitions collected by an unknown behavior policy, without further environment interaction.

Our goal is to learn each agent's policy as a decentralized actor through centralized training.
During training, we first use a centralized generative teacher that has access to the joint context $\mathbf{x}$. The teacher is represented by a deterministic sampler conditioned on source noise distribution:
\begin{equation}
\tilde{\mathbf a}=\mu_\psi(\mathbf{x},\mathbf{z}),
\qquad
\mathbf{z}=(z_1,\ldots,z_N)\sim\rho_0,
\label{eq:centralized_teacher_sampler}
\end{equation}
where \(\rho_0\) is a simple source noise distribution, such as a standard Gaussian.
Since $\mathbf{z}$ is random, this sampler induces a conditional distribution over
joint actions given $\mathbf{x}$. In the teacher stage, both the full joint observation context
$\mathbf{x}$ and the full joint noise $\mathbf{z}$ may be used.
At execution time, however, each agent must act using only its local
information. We therefore deploy a strict-product policy:
\begin{equation}
\pi_\mathbf{w}^{\mathrm{prod}}(\mathbf a\mid \mathbf x)
=
\prod_{i=1}^N \pi_{w_i}(a_i\mid o_i),
\label{eq:strict_product}
\end{equation}
where \(\mathbf{w}=(w_1,\ldots,w_N)\) denotes the set of decentralized actor parameters across $N$ agents. In the
one-step reparameterized implementation used throughout this work, the deployed
action is
\(\hat{\mathbf a}=(\mu_{w_1}(o_1,z_1),\ldots,\mu_{w_N}(o_N,z_N))\),
with each actor receiving only its local observation \(o_i\) and local noise
block \(z_i\), unless specified otherwise. Thus, the centralized teacher can
model coordinated joint behavior using $\mathbf{x}$ and $\mathbf{z}$, while the student must reproduce such behavior under this decentralized information constraint.

For value-guided actor extraction, we use a centralized
critic $Q_{\text{tot}}(\mathbf{x},\mathbf{a})$ that satisfies the Individual-Global-Max (IGM)
principle~\citep{sunehag2017value, rashid2020monotonic, son2019qtran}:
\begin{equation}
\arg\max_{\mathbf{a}}\, Q_{\text{tot}}(\mathbf{x},\mathbf{a})
\;=\;\Bigl(\arg\max_{a_1} Q_1(o_1,a_1),\,\ldots,\,\arg\max_{a_N} Q_N(o_N,a_N)\Bigr).
\label{eq:igm}
\end{equation}
Here, $Q_{\text{tot}}(\mathbf{x},\mathbf{a})$ denotes the global Q-function over the joint context and
joint action, while $Q_i(o_i,a_i)$ denotes the individual utility for agent~$i$.
The IGM principle ensures that greedy selection over each local $Q_i$ remains
consistent with the global optimum of $Q_{\text{tot}}$.

The remainder of Sec.~\ref{sec:method} describes how we train the teacher so that its source
noise is assigned to coordinated joint-action modes, and how we distill the
resulting teacher into decentralized actors.

\begin{figure}[t]
  \centering
  \includegraphics[width=0.99\linewidth]{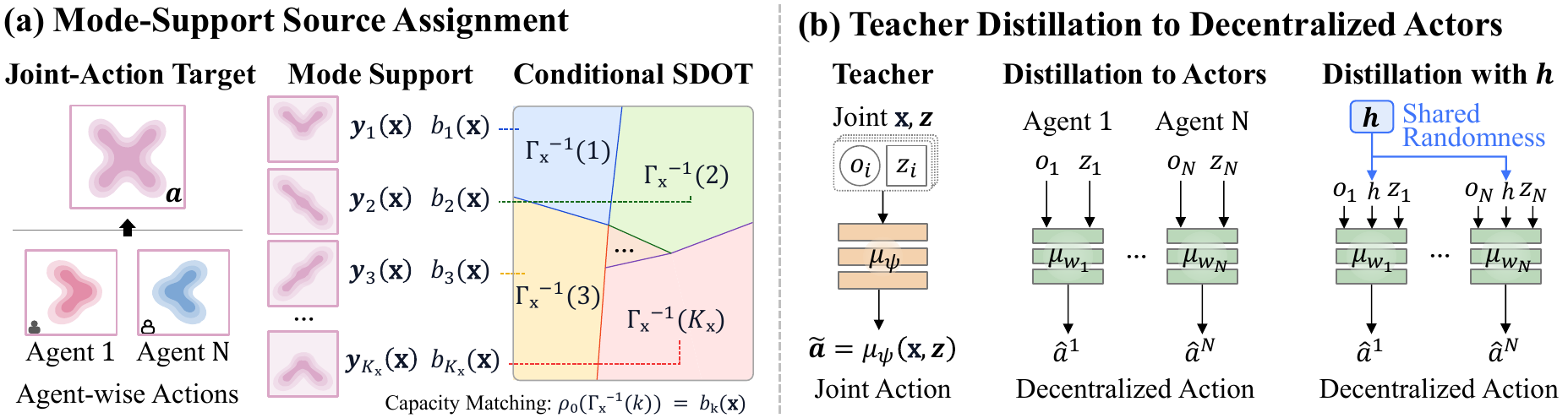}
  \caption{\textbf{MoSDOT-based framework for multi-agent distillation.}
\textbf{(a)} Replay joint actions are summarized into a mode support $\{\mathbf{y}_k(\mathbf{x})\}$ with capacities $\{b_k(\mathbf{x})\}$, and conditional SDOT partitions the noise space into Laguerre cells $\Gamma_\textbf{x}^{-1}(k)$ assigned to each mode. \textbf{(b)} The centralized teacher $\mu_\psi$ is distilled into decentralized one-step actors $\{\mu_{w_i}\}$, each using only its local observation and noise. An optional shared signal $h$ correlates local noise to mitigate the strict-product gap.}
  \label{fig:framework}
  \vspace{-\baselineskip}
\end{figure}

\subsection{Mode Support and Capacities}
\label{subsec:mode_support}
Learning the teacher sampler \(\mu_\psi(\mathbf{x},\mathbf{z})\) requires pairing source noise \(\mathbf{z}\) with target joint actions so that the induced conditional distribution matches the target behavior at \(\mathbf{x}\).
A naive approach pairs \(\mathbf{z}\) and the replay target without considering their relationship, mapping them arbitrarily.
When the targets at a given \(\mathbf{x}\) contain multiple distinct coordinated joint behaviors, such as two valid joint actions in a coordination game, random pairing entangles these behaviors in source-noise space.
This often causes nearby noise vectors to map to unrelated joint actions, and the noise mass allocated to each behavior is determined by random sampling rather than designed intentionally.

We instead organize the targets at each $\mathbf{x}$ into a finite \emph{mode support}, a discrete set of representative joint actions, and assign a \emph{capacity} (a target source-noise mass) to each element of this support before training.
The sampler then learns a structured assignment in which contiguous regions of $\mathbf{z}$-space correspond to single modes, with region sizes set by the capacities.
\vspace{-3pt}
\paragraph{Mode support.}
For each conditioning context \(\mathbf{x}\), we collect a finite set of \(M_\textbf{x}\) target joint actions:
\begin{equation}
\mathcal T_\textbf{x}=\{\mathbf{a}^{(m)}(\mathbf{x})\}_{m=1}^{M_\textbf{x}},
\qquad
\mathbf{a}^{(m)}(\mathbf{x})\in\mathcal A_1\times\cdots\times\mathcal A_N .
\label{eq:target_joint_action_set}
\end{equation}
In the default construction, these targets are replay actions from datasets.
A support aggregation rule
\begin{equation}
q_\textbf{x}:\{1,\ldots,M_\textbf{x}\}\rightarrow\{1,\ldots,K_\textbf{x}\}
\label{eq:support_aggregation_rule}
\end{equation}
groups the targets into \(K_\textbf{x}\) clusters. Each cluster \(k\) collects the target indices \(C_k(\mathbf{x})=\{m:q_\textbf{x}(m)=k\}\) and is summarized by a representative joint action \(\mathbf{y}_k(\mathbf{x})\in\mathcal A_1\times\cdots\times\mathcal A_N\), which we call a \emph{mode}. The \emph{mode support} at context \(\mathbf{x}\) is the discrete set of these representatives:
\begin{equation}
\mathcal Y_\textbf{x}=\{\mathbf{y}_1(\mathbf{x}),\ldots,\mathbf{y}_{K_\textbf{x}}(\mathbf{x})\}.
\label{eq:mode_support_set}
\end{equation}
The support size \(K_\textbf{x}\) is determined by the observed targets and the aggregation rule, not by enumerating the full combinatorial joint action space.
The aggregation rule $q_\textbf{x}$ determines the granularity of $\mathcal Y_\textbf{x}$.
We use identity support for discrete-action benchmarks and a simple joint-action quantization for continuous-action benchmarks.

\vspace{-3pt}
\paragraph{Capacities.}
Each mode \(\mathbf{y}_k(\mathbf{x})\) is assigned a \emph{capacity} \(b_k(\mathbf{x})\ge 0\) satisfying \(\sum_{k=1}^{K_\textbf{x}} b_k(\mathbf{x})=1\), specifying its target source-noise mass.
We use \(b_k(\mathbf{x})=|C_k(\mathbf{x})|/M_\textbf{x}\) so that the target source-noise mass matches the empirical replay frequency of each mode.
Although each mode decomposes as $\mathbf{y}_k(\mathbf{x})=\bigl(y_{k,1}(\mathbf{x}),\ldots,y_{k,N}(\mathbf{x})\bigr)$, source assignment is performed on $\mathbf{y}_k(\mathbf{x})$ as a whole, so source-noise regions are assigned to coordinated joint behaviors rather than to a product of per-agent action modes.
This is what allows the sampler to represent correlated joint behaviors that a per-agent factorization cannot.

\subsection{Source Assignment via Semi-Discrete Optimal Transport}
\label{subsec:teacher_source_assignment}
Sec.~\ref{subsec:mode_support} defines the target mass \(b_k(\textbf{x})\) for each mode, but leaves open which source-noise samples should be paired with which mode.
We resolve this by partitioning source-noise space so that each cell carries the prescribed mass and is assigned to a single mode.
We adopt the SDOT-based source--target alignment introduced for image generation by~\citet{kong2025alignflowimproving}, and apply it to the capacity-weighted mode support \(\mathcal Y_\textbf{x}\).

Let \(\rho_0\) denote the source law over joint noise \(\mathbf{z}=(z_1,\ldots,z_N)\). We define the discrete target measure at context \(\mathbf{x}\) as follows:
\begin{equation}
\nu_\textbf{x}
=
\sum_{k=1}^{K_\textbf{x}} b_k(\textbf{x})\,\delta_{\mathbf{y}_k(\mathbf{x})},
\label{eq:mode_target_measure}
\end{equation}
where \(\delta_{\mathbf{y}_k(\mathbf{x})}\) denotes a Dirac measure at mode \(\mathbf{y}_k(\mathbf{x})\), so \(\nu_\textbf{x}\) places mass \(b_k(\textbf{x})\) on each mode in \(\mathcal Y_\textbf{x}\).
For a cost \(c_\textbf{x}(\mathbf{z},k)\) between source noise \(\mathbf{z}\) and mode \(\mathbf{y}_k(\mathbf{x})\), the SDOT primal is defined as follows:
\begin{equation}
\min_{\Gamma_\textbf{x}}
\mathbb E_{\mathbf{z}\sim\rho_0}
\!\left[
c_\textbf{x}\bigl(\mathbf{z},\Gamma_\textbf{x}(\mathbf{z})\bigr)
\right]
\quad
\mathrm{s.t.}
\quad
\rho_0\!\left(\Gamma_\textbf{x}^{-1}(k)\right)=b_k(\textbf{x}),
\quad
k=1,\ldots,K_\textbf{x} .
\label{eq:mosdot_primal}
\end{equation}
Here, \(\Gamma_\textbf{x}:\mathbf{z}\mapsto k\) is the source-to-mode assignment, mapping each source sample to a mode index.
We take \(c_\textbf{x}\) to be the squared Euclidean distance in the representation that defines \(\mathcal Y_\textbf{x}\).
The SDOT dual is a finite-dimensional concave problem and admits stochastic optimization on samples from \(\rho_0\)~\citep{kitagawa2019convergence}. We write its objective as:
\begin{equation}
\Phi_\textbf{x}(g)
=
\sum_{k=1}^{K_\textbf{x}} b_k(\textbf{x})g_k
+
\mathbb E_{\mathbf{z}\sim\rho_0}
\!\left[
\min_{j\in\{1,\ldots,K_\textbf{x}\}}
\bigl(c_\textbf{x}(\mathbf{z},j)-g_j\bigr)
\right].
\label{eq:sdot_dual_objective}
\end{equation}
The optimal dual weights are obtained as \(g_\textbf{x}^\star\in\operatorname*{arg\,max}_{g\in\mathbb R^{K_\textbf{x}}}\Phi_\textbf{x}(g)\).
The resulting Laguerre cells define the source-to-mode assignment:
\begin{equation}
\Gamma_\textbf{x}(\mathbf{z})
=
\operatorname*{arg\,min}_{k}
\bigl[
c_\textbf{x}(\mathbf{z},k)-g_{x,k}^\star
\bigr].
\label{eq:source_to_mode_assignment}
\end{equation}
We solve Eq.~\ref{eq:sdot_dual_objective} stochastically with Monte-Carlo sampling from \(\rho_0\).
For each pair \((\mathbf{x},\mathbf{z})\), the assignment \(\Gamma_\textbf{x}\) selects the mode index, and we write the corresponding mode as \(\mathbf{y}_\Gamma(\mathbf{x},\mathbf{z})=\mathbf{y}_{\Gamma_\textbf{x}(\mathbf{z})}(\mathbf{x})\).
We then form the linear interpolation path \(\bar{\mathbf{a}}_t=(1-t)\mathbf{z}+t\,\mathbf{y}_\Gamma(\mathbf{x},\mathbf{z})\) with displacement \(\mathbf{v}_\Gamma(\mathbf{x},\mathbf{z})=\mathbf{y}_\Gamma(\mathbf{x},\mathbf{z})-\mathbf{z}\), where \(t\sim\mathrm{Unif}([0,1])\).
Following~\citep{lipman2022flow}, the teacher learns a velocity field \(u_\psi\) by regressing onto the assigned displacement:
\begin{equation}
\mathcal L_{\mathrm{teacher}}(\psi)
=
\mathbb E_{\mathbf{x}\sim\mathcal D,\,\mathbf{z}\sim\rho_0,\,t\sim\mathrm{Unif}([0,1])}
\!\left[
\left\|
u_\psi(\bar{\mathbf{a}}_t,t,\mathbf{x})
-
\mathbf{v}_\Gamma(\mathbf{x},\mathbf{z})
\right\|_2^2
\right].
\label{eq:assignment_teacher_loss}
\end{equation}
At inference, starting from \(\mathbf{z}\) and integrating the learned field \(u_\psi\) from \(t=0\) to \(t=1\) defines the teacher sampler \(\mu_\psi(\mathbf{x},\mathbf{z})\) used in Sec.~\ref{subsec:problem_setup}.
Unlike independent pairing, this couples source regions to modes with capacity-controlled mass.

\subsection{Joint Policy Distillation}
\label{subsec:joint_policy_distillation}
The teacher trained in Sec.~\ref{subsec:teacher_source_assignment} produces coordinated joint-action samples using the full context and joint noise.
We distill these rollouts into the strict-product student family in Eq.~\eqref{eq:strict_product}.
For a replay context \(\mathbf{x}=(o_1,\ldots,o_N)\) and source noise \(\mathbf{z}=(z_1,\ldots,z_N)\sim\rho_0\), the teacher and student actions are defined as:
\begin{equation}
\tilde{\mathbf{a}}
=
\mu_\psi(\mathbf{x},\mathbf{z}),
\qquad
\hat{\mathbf{a}}
=
\bigl(
\mu_{w_1}(o_1,z_1),
\ldots,
\mu_{w_N}(o_N,z_N)
\bigr).
\label{eq:teacher_student_rollouts}
\end{equation}
The distillation loss matches each local actor to the corresponding component of the centralized teacher rollout:
\begin{equation}
\mathcal L_{\mathrm{distill}}(\mathbf{w})
=
\mathbb E_{\mathbf{x}\sim\mathcal D,\,\mathbf{z}\sim\rho_0}
\left[
\sum_{i=1}^N
\left\|
\mu_{w_i}(o_i,z_i)
-
[\mu_\psi(\mathbf{x},\mathbf{z})]_i
\right\|_2^2
\right],
\label{eq:distillation_loss}
\end{equation}
where \([\cdot]_i\) extracts the action component for agent \(i\).
For value-guided actor extraction, we optimize:
\begin{equation}
\mathcal L_{\pi}(\mathbf{w})
=
\mathbb E_{\mathbf{x}\sim\mathcal D,\,\mathbf{z}\sim\rho_0}
\left[
-
Q_{\mathrm{tot}}(\mathbf{x},\hat{\mathbf{a}})
+
\alpha
\sum_{i=1}^N
\left\|
\mu_{w_i}(o_i,z_i)
-
[\mu_\psi(\mathbf{x},\mathbf{z})]_i
\right\|_2^2
\right],
\label{eq:value_guided_actor_loss}
\end{equation}
where $\hat{\mathbf{a}}$ is as defined in Eq.~\eqref{eq:teacher_student_rollouts}, $Q_{\mathrm{tot}}$ is the IGM-compatible centralized critic of Eq.~\eqref{eq:igm}, and $\alpha$ controls the strength of teacher regularization.
The assignment \(\Gamma_\textbf{x}\) is used only to train the teacher and is not given to the decentralized actors at execution time.

\paragraph{Strict-product gap.}
MoSDOT improves the centralized teacher, but it does not remove the execution-time factorization constraint.
A strict-product actor cannot represent every correlated joint distribution.
For example, even when MoSDOT places clean teacher mass on \((+1,-1)\) and \((-1,+1)\), a strict-product student with the same local marginals also assigns mass to unsupported tuples \((+1,+1)\) and \((-1,-1)\): each agent independently samples its own noise, so the two cannot agree on which mode to execute.
We call this residual mismatch the \emph{strict-product gap}.

The MoSDOT construction itself suggests a way to probe this gap.
Recall that the assignment \(\Gamma_\textbf{x}\) partitions the noise space into Laguerre cells, each mapped to a single mode \(\mathbf{y}_k(\mathbf{x})\).
If all agents observe the same cell index, they automatically agree on the same mode and avoid unsupported product tuples.
This motivates a shared-randomness variant.
Let \(h\sim\rho_h\) be a random signal shared across agents at execution, playing the role of a shared cell index, and let \(\epsilon_i\sim\rho_i\) be agent-local noise for within-mode variability.
The resulting policy can be modified as:
\begin{equation}
\pi_{\mathbf{w}}^{\mathrm{sr}}(\mathbf{a}\mid \mathbf{x})
=
\mathbb E_{h\sim\rho_h}
\left[
\prod_{i=1}^N
\pi_{w_i}^{\mathrm{sr}}(a_i\mid o_i,h)
\right],
\label{eq:shared_randomness_policy}
\end{equation}
with one-step implementation
\(\hat{\mathbf{a}}^{\mathrm{sr}}=
\bigl(\mu_{w_1}^{\mathrm{sr}}(o_1,h,\epsilon_1),\ldots,
\mu_{w_N}^{\mathrm{sr}}(o_N,h,\epsilon_N)\bigr)\).
Without \(h\), this reduces to strict-product execution as in Eq.~\ref{eq:strict_product}.
With \(h\), agents act locally and exchange no observations, actions, or messages, but their mode choice is correlated through the shared signal.
This isolates the residual error due to independent local sampling and complements the teacher-side coupling provided by MoSDOT.
Further theoretical and algorithmic details are provided in
Appendices~\ref{app:shared_randomness_closes_gap} and~\ref{app:alg_mosdot_sr}.

\vspace{-4pt}
\section{Experiments}
\label{sec:experiments}
\vspace{-4pt}

Our experiments separate three sources of error in CTDE generative
distillation: (i) teacher-side off-support mass, (ii) incompatibility
between the teacher's source assignment and decentralized actors, and
(iii) the residual strict-product gap. MoSDOT is designed to address first two by
replacing independent source--target pairing with a capacity-matched SDOT
assignment. We then compare strict-product and shared-randomness decentralized
actors to isolate what can be preserved under decentralized execution.

\vspace{-5pt}
\subsection{Experimental Setup}
\label{subsec:exp_setup}

\begin{figure}[t]
  \centering
  \includegraphics[width=0.8\linewidth]{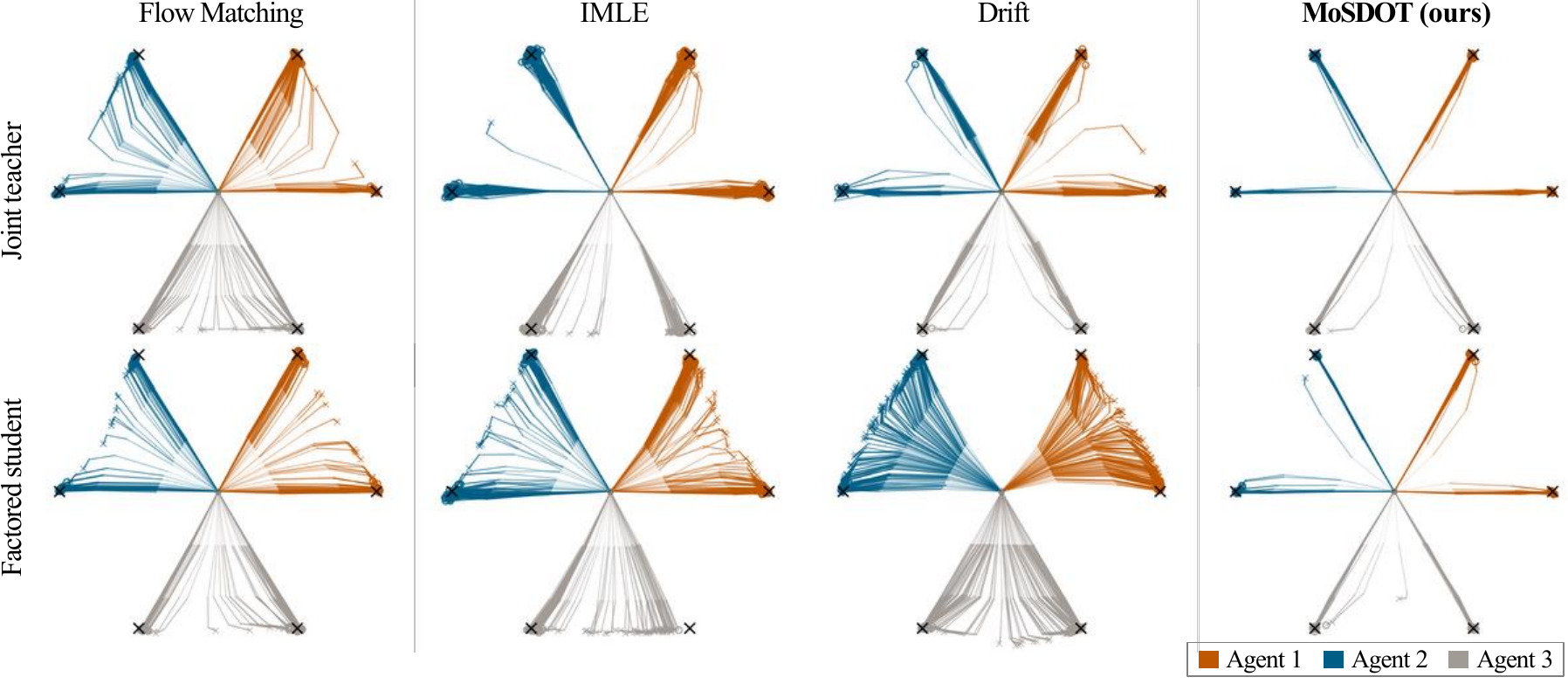}
  \vspace{-5pt}
  \caption{\textbf{Landmark policy trajectories.}
    Black $\times$ marks denote the six anchor mode endpoints, and colors indicate
    the three anchor pairs. The top row shows centralized joint teacher rollouts,
    and the bottom row shows one-step factorized students distilled under
    independent local noise. Baseline students show fan mass between modes, while
    MoSDOT reduces fan mass and retains the structure.}
  \label{fig:landmark_trajectories}
  \vspace{-\baselineskip}
\end{figure}

\vspace{-4pt}
\paragraph{Diagnostics and benchmarks.}
We use two controlled tasks to visualize the failure modes studied in this
work. The landmark diagnostic in Fig.~\ref{fig:landmark_trajectories} evaluates
teacher-side source assignment with multiple coordinated modes across three
agents. The XOR task in Fig.~\ref{fig:xor_strict_product}, adapted from
MAC-Flow~\citep{lee2026macflow}, isolates the strict-product gap using the
anti-correlated support $\{(-1,+1),(+1,-1)\}$. For benchmark evaluation, we use
SMACv1~\citep{samvelyan2019starcraft}, SMACv2~\citep{ellis2023smacv2}, and MPE~\citep{lowe2017multi}.

\vspace{-4pt}
\paragraph{Baselines and protocol.}
For controlled tasks, we compare against independent flow
matching~\citep{lipman2022flow}, IMLE policy~\citep{rana2025imlepolicy}, and
Drifting model~\citep{deng2026drifting}. For benchmark evaluation, we follow
the protocol and baseline suite of MAC-Flow~\citep{lee2026macflow}, which
includes representative Gaussian, diffusion, and flow-policy baselines such as
MADiff, DoF, and MAC-Flow~\citep{zhu2024madiff,li2025dof,lee2026macflow}.
Unless stated otherwise, benchmark results use \(6\) seeds and report two
standard deviations after \(\pm\). More details on baselines and implementations are provided in Appendix~\ref{app:exp_details}.

\paragraph{Diagnostic metrics.}
For the landmark diagnostic, we evaluate teacher quality along two axes.
\textbf{(i) Endpoint support}: does the teacher place samples on the modes
in \(\mathcal Y_\textbf{x}\)? \textit{Fan-free} counts actions inside accepted
directional cones. \textit{Success} measures landmark reach rate, and
\textit{Balance} measures balance between valid local modes.
\textbf{(ii) Source routing}: how does the teacher use source noise to
choose among modes? \textit{Route consistency} combines fan-free mass with
the stability of mode labels among nearby samples in an agent's noise slice.
\textit{Jacobian off-diagonal ratio} measures how strongly the teacher's
output for agent~$i$ depends on other agents' noise $z_{j\neq i}$ relative
to its own noise $z_i$, with values near zero indicating weak local
cross-noise sensitivity. Endpoint metrics test
whether the teacher reaches \(\mathcal Y_\textbf{x}\), and routing metrics test
whether mode choices depend only on locally observable information.
Full definitions are given in Appendix~\ref{app:metric_defs}.

\begin{table}[t]
\setlength{\fboxsep}{1pt}
\centering
\begin{minipage}[t]{0.52\linewidth}
  \vspace{0pt}  %
  \centering
  \footnotesize
  \setlength{\tabcolsep}{2pt}
  \renewcommand{\arraystretch}{1.3}
  \newcommand{\landmarkscore}[2]{\mbox{{#1}\,{\fontsize{6}{7}\selectfont$\pm#2$}}}
  \newcommand{\landmarkbest}[2]{\cellcolor{tblBest}\landmarkscore{\textbf{#1}}{#2}}
  \newcommand{\landmarksecond}[2]{\cellcolor{tblSecond}\landmarkscore{\underline{#1}}{#2}}
  \resizebox{\linewidth}{!}{%
  \begin{tabular}{@{}lcccc@{}}
    \toprule
    Method & Fan-free $\uparrow$ & Route cons.\ $\uparrow$ & Success $\uparrow$ & Balance $\uparrow$ \\
    \midrule
    Flow matching
      & \landmarkscore{0.857}{0.010} & \landmarkscore{0.856}{0.010}
      & \landmarkscore{0.915}{0.029} & \landmarkscore{0.917}{0.043} \\
    IMLE
      & \landmarkbest{0.991}{0.007} & \landmarksecond{0.857}{0.023}
      & \landmarkscore{0.848}{0.064} & \landmarkscore{0.761}{0.029} \\
    Drifting
      & \landmarkscore{0.952}{0.005} & \landmarkscore{0.653}{0.023}
      & \landmarksecond{0.985}{0.002} & \landmarksecond{0.936}{0.029} \\
    \textbf{MoSDOT}
      & \landmarksecond{0.983}{0.002} & \landmarkbest{0.972}{0.001}
      & \landmarkbest{0.991}{0.002} & \landmarkbest{0.961}{0.011} \\
    \bottomrule
  \end{tabular}%
  }
  \vspace{4pt}
  \captionsetup{font=small}
  \caption{\textbf{Joint-teacher landmark diagnostics.} Mean $\pm$ one standard deviation over 4 seeds.}
  \label{tab:landmark_endpoint}
\end{minipage}%
\hfill
\begin{minipage}[t]{0.46\linewidth}
  \vspace{0pt}
  \centering
  \includegraphics[width=1.0\linewidth, trim={0 10pt 0 0}, clip]{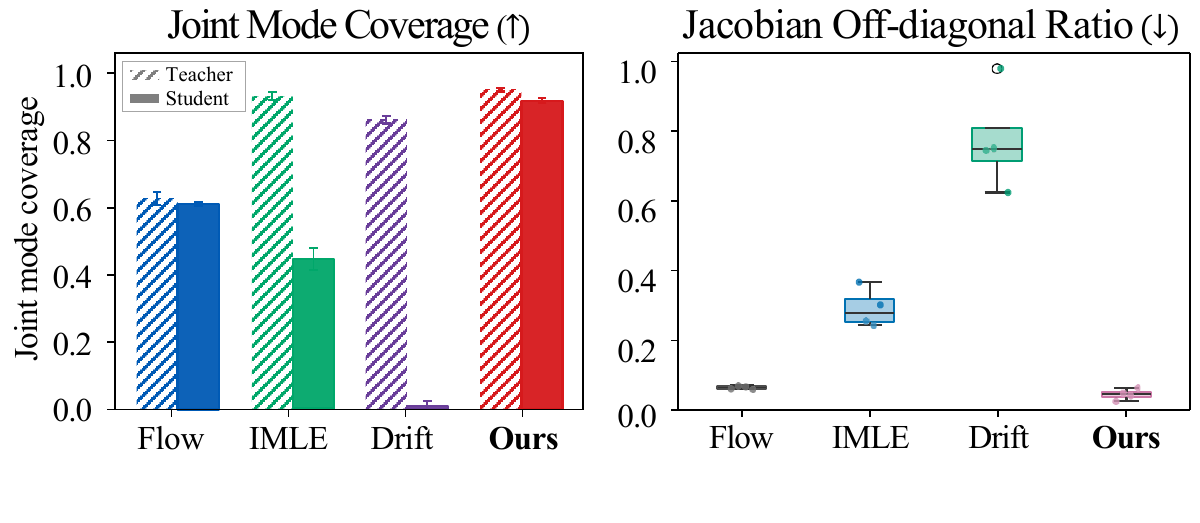}
  \captionsetup{font=small}
  \captionof{figure}{\textbf{Landmark route-locality diagnostics.}}
  \label{fig:route_locality}
\end{minipage}
\vspace{-2\baselineskip}
\end{table}

\subsection{Endpoint Accuracy Does Not Imply Distillable Assignments}
\label{subsec:rq1}
We first test whether reaching the correct modes is sufficient for factorized
distillation. Tab.~\ref{tab:landmark_endpoint} measures endpoint-level support
quality, while Fig.~\ref{fig:route_locality} asks a complementary question
about whether the teacher's mode choice for an agent depends on the agent's
own noise or the noise of other agents.

A teacher can reach valid modes while choosing among them in a way
unavailable to the corresponding local actor. Flow matching has reasonable
route consistency and balance, but its lower fan-free score reveals samples
between modes. IMLE achieves the best fan-free score with sharp endpoints,
but sacrifices success and balance, appearing as uneven mass across modes
in Fig.~\ref{fig:route_locality}. Drifting shows this mismatch most
strongly. It reaches valid modes often, but has the lowest route
consistency in Tab.~\ref{tab:landmark_endpoint} and the largest Jacobian
off-diagonal ratio in Fig.~\ref{fig:route_locality}, indicating its
per-agent output is strongly shaped by other agents' noise—information
unavailable to the local actor.
Endpoint support is therefore insufficient: when mode choice depends on
off-agent noise, distillation can average or mix modes even with accurate
endpoints. We evaluate teacher quality along both axes—whether it reaches
\(\mathcal Y_\textbf{x}\), and whether mode choices are reproducible from locally
available information.

\begin{wrapfigure}[14]{r}{0.3\linewidth}
  \renewcommand{\thefigure}{\number\numexpr\value{figure}+1\relax}
  \vspace{-20pt}  %
  \centering
  \includegraphics[width=\linewidth]{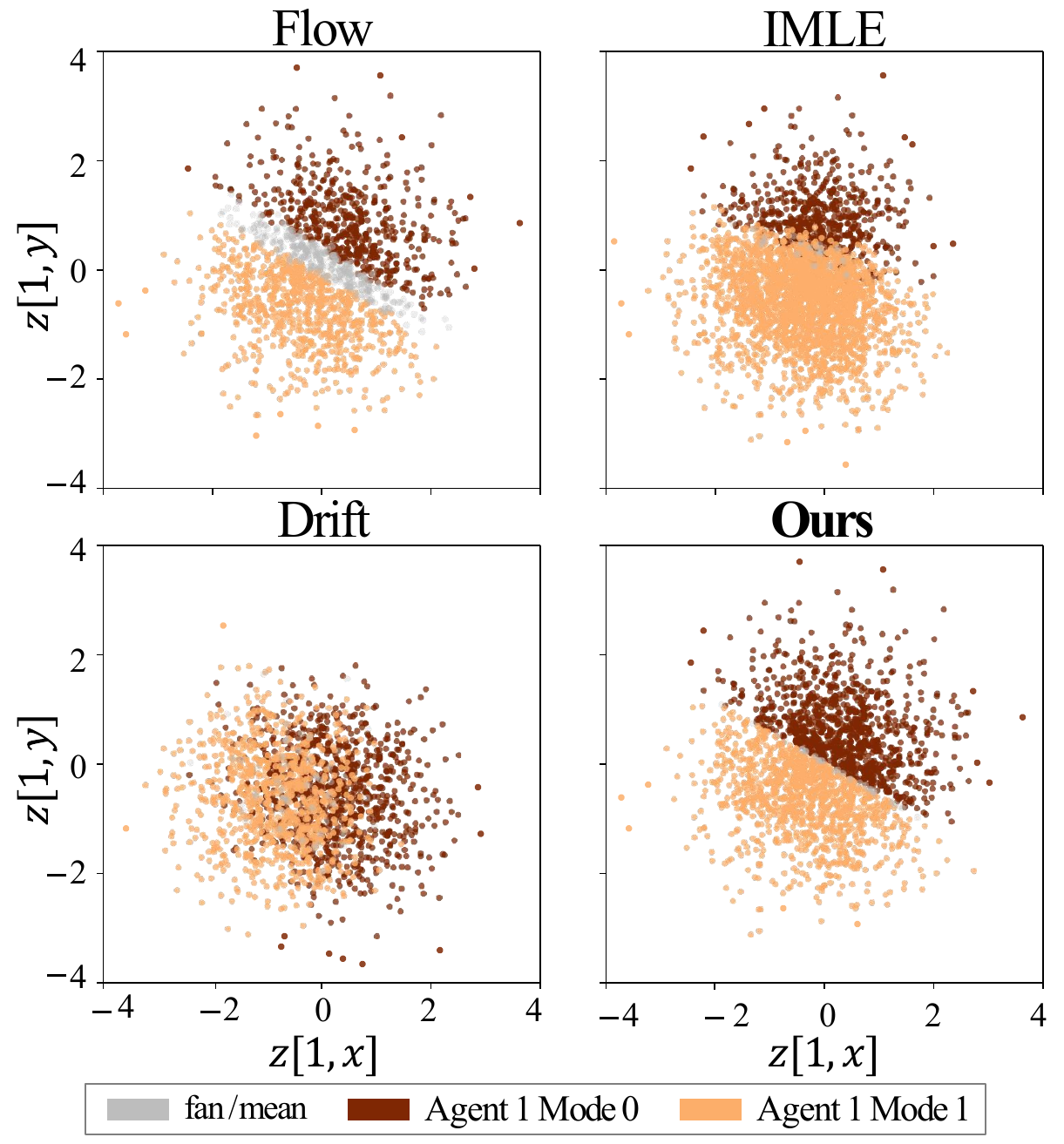}
  \vspace{-\baselineskip}
  \caption{Source-space partitions in Agent~1's noise plane.}
  \vspace{-15pt}
  \label{fig:conditional_partition_2x2}
\end{wrapfigure}

\vspace{-4pt}
\subsection{MoSDOT Preserves Support Without Entangling Assignments}
\label{subsec:rq2}
The previous section showed that no baseline is strong on both axes
simultaneously. We now evaluate whether MoSDOT closes this gap. On the
same landmark replay, MoSDOT replaces independent source--target pairing
with the capacity-matched SDOT assignment of
Eq.~\eqref{eq:assignment_teacher_loss}, using replay frequencies as mode
capacities.

MoSDOT improves support sharpness without sacrificing assignment locality.
On the endpoint axis (Tab.~\ref{tab:landmark_endpoint}), it is competitive
with the best endpoint-focused baselines on fan-free support and success,
and achieves the best balance. The teacher's joint mode coverage in
Fig.~\ref{fig:route_locality} (left, hatched) is also the highest across
methods. On the routing axis, its route consistency is the highest
among all methods, and its Jacobian off-diagonal ratio in
Fig.~\ref{fig:route_locality} (right) is comparable to flow matching, well
below IMLE and Drifting.
Fig.~\ref{fig:conditional_partition_2x2} visualizes both axes jointly in
Agent~1's noise plane: Flow leaves a fan between modes, IMLE assigns
unbalanced area, and Drift mixes the two modes throughout, while MoSDOT
yields balanced, fan-less, disentangled regions.

\vspace{-4pt}
\subsection{Cleaner Source Assignments Survive Decentralized Distillation}
\label{subsec:rq3}
Teacher-side improvement is useful only if it persists after decentralized
distillation. To examine the teacher-to-student transfer, we distill
strict-product students from Flow Matching, IMLE, Drifting, and MoSDOT teachers
under the same student architecture, seed list, aggregation rule, and training
budget. MoSDOT uses the capacity-matched teacher from Sec.~\ref{subsec:rq2}.

The student appears to track the teacher's geometry.
Fig.~\ref{fig:landmark_trajectories} (bottom row) shows that the MoSDOT
student largely follows the teacher's ray structure, while baselines show
fan mass between modes. Quantitatively, the MoSDOT-distilled student
retains higher fan-free, success, and balance scores than baselines, while
IMLE and Drifting students lose substantial endpoint quality through distillation
(Appendix~\ref{app:student_diagnostics}). These results suggest that
improving the centralized source assignment can improve the support
preserved by decentralized actors.

\begin{figure}[t]
  \centering
  \includegraphics[width=0.8\linewidth]{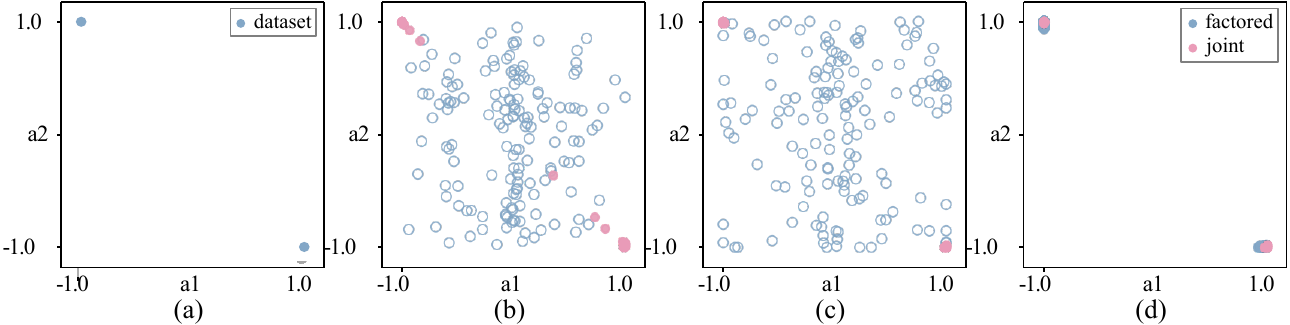}
  \renewcommand{\thefigure}{\number\numexpr\value{figure}-1\relax}
  \caption{
    \textbf{XOR strict-product visualization.}
    \textbf{(a)} Target support placing equal mass on $(-1,+1)$ and $(+1,-1)$.
    \textbf{(b)} \textbf{MAC-Flow}: bridge mass spreads across all quadrants under strict-product execution.
    \textbf{(c)} \textbf{Ours w/o shared signal}: clean teacher, residual strict-product gap.
    \textbf{(d)} \textbf{Ours with shared signal}: gap closed.
    Blue: factored student; pink: joint teacher.
    }
  \label{fig:xor_strict_product}
  \vspace{-\baselineskip}
\end{figure}

\vspace{-4pt}
\subsection{Shared Randomness Isolates the Strict-Product Gap}
\label{subsec:rq4}
Even with a clean teacher, the strict-product student of
Eq.~(\ref{eq:strict_product}) cannot in general represent correlated joint
support. The XOR target $\{(-1,+1),(+1,-1)\}$ illustrates this: independent
local sampling that matches the marginals also produces the unsupported
tuples $(+1,+1)$ and $(-1,-1)$. We use the shared-randomness variant of
Eq.~(\ref{eq:shared_randomness_policy}) to separate this product-projection
error from teacher-side assignment error.
Fig.~\ref{fig:xor_strict_product} shows the separation across three stages.
The unrouted flow teacher places bridge mass between the two valid tuples,
which a strict-product student spreads across all four quadrants (b).
MoSDOT concentrates the teacher mass on the two valid tuples, yet
independent local sampling still mixes the two modes after distillation
(c). Shared randomness, with the same MoSDOT teacher, lets the actors
agree on a mode-selection signal while still acting locally, and the
student concentrates on the valid tuples (d). The transition from (c) to
(d)—where only the execution-time signal changes—isolates the strict-product
gap from teacher-side assignment failure.

\begin{table}[t]
\centering
\setlength{\fboxsep}{1pt}
\caption{\textbf{Performance on discrete-action offline MARL benchmarks
(SMACv1 and SMACv2).} Mean over $6$ seeds, $\pm 2\sigma$. Best and
second-best in \colorbox{tblBest}{\strut sage} and
\colorbox{tblSecond}{\strut powder blue}.}
\label{tab:smac_perf}
\scriptsize
\setlength{\tabcolsep}{2.4pt}
\renewcommand{\arraystretch}{1.08}
\resizebox{0.95\textwidth}{!}{
\begin{tabular}{l l l c c c c c c c c c}
\toprule
\multirow{2}{*}{\textbf{Benchmark}}
& \multirow{2}{*}{\textbf{Scenario}}
& \multirow{2}{*}{\textbf{Dataset}}
& \multicolumn{3}{c}{\texttt{Offline / BC baselines}}
& \multicolumn{3}{c}{\texttt{Diffusion policies}}
& \multicolumn{3}{c}{\texttt{Flow / OT policies}} \\
\cmidrule(lr){4-6} \cmidrule(lr){7-9} \cmidrule(lr){10-12}
& &
& \texttt{BC}
& \texttt{MABCQ}
& \texttt{MACQL}
& \texttt{Diffusion BC}
& \texttt{MADiff}
& \texttt{DoF}
& \texttt{Flow BC}
& \texttt{MAC-Flow}
& \texttt{MoSDOT} \\
\midrule

{\multirow{16}{*}{\STAB{\rotatebox[origin=c]{90}{\textbf{SMACv1}}}}}
& \multirow{3}{*}{\textbf{3m}}
& Good
& \score{16.0}{1.0}
& \score{3.7}{1.1}
& \score{19.1}{0.1}
& \score{19.5}{0.5}
& \score{19.3}{0.5}
& \second{\score{19.8}{0.2}}
& \best{\score{20.0}{0.0}}
& \second{\score{19.8}{0.2}}
& \best{\score{20.0}{0.0}} \\
& & Medium
& \score{8.2}{0.8}
& \score{4.0}{1.0}
& \score{13.7}{0.3}
& \score{13.3}{0.7}
& \score{16.4}{2.6}
& \second{\score{18.6}{1.2}}
& \score{14.7}{1.5}
& \score{18.0}{3.2}
& \best{\score{18.8}{1.5}} \\
& & Poor
& \score{4.4}{0.1}
& \score{3.4}{1.0}
& \score{4.2}{0.1}
& \score{4.2}{0.2}
& \score{10.3}{6.1}
& \second{\score{10.9}{1.1}}
& \score{4.5}{0.1}
& \score{10.6}{2.2}
& \best{\score{16.4}{4.9}} \\

\cmidrule(lr){2-3} \cmidrule(lr){4-12}

& \multirow{3}{*}{\textbf{8m}}
& Good
& \score{16.7}{0.4}
& \score{4.8}{0.6}
& \score{18.9}{0.9}
& \score{19.4}{0.5}
& \score{18.9}{1.1}
& \score{19.6}{0.3}
& \score{19.5}{0.2}
& \second{\score{19.7}{0.3}}
& \best{\score{20.0}{0.0}} \\
& & Medium
& \score{10.7}{0.5}
& \score{5.6}{0.6}
& \score{15.5}{1.5}
& \score{18.6}{0.6}
& \score{16.8}{1.6}
& \score{18.6}{0.8}
& \score{18.2}{0.8}
& \second{\score{19.4}{0.6}}
& \best{\score{20.0}{0.0}} \\
& & Poor
& \score{5.3}{0.1}
& \score{3.6}{0.8}
& \score{7.5}{1.0}
& \score{4.8}{0.2}
& \score{9.8}{0.9}
& \best{\score{12.0}{1.2}}
& \score{4.9}{0.1}
& \second{\score{11.5}{0.8}}
& \score{10.8}{0.8} \\

\cmidrule(lr){2-3} \cmidrule(lr){4-12}

& \multirow{3}{*}{\textbf{2s3z}}
& Good
& \score{18.2}{0.4}
& \score{7.7}{0.9}
& \score{17.4}{0.3}
& \score{18.0}{1.0}
& \score{15.9}{1.2}
& \score{18.5}{0.8}
& \second{\score{19.5}{0.1}}
& \second{\score{19.5}{0.5}}
& \best{\score{20.1}{0.1}} \\
& & Medium
& \score{12.3}{0.7}
& \score{7.6}{0.7}
& \score{15.6}{0.4}
& \score{13.4}{1.4}
& \score{15.6}{0.3}
& \second{\score{18.1}{0.9}}
& \score{15.1}{2.0}
& \score{17.6}{0.6}
& \best{\score{18.2}{1.0}} \\
& & Poor
& \score{6.7}{0.3}
& \score{6.6}{0.2}
& \score{8.4}{0.8}
& \score{6.2}{1.2}
& \score{8.5}{1.3}
& \best{\score{10.0}{1.1}}
& \score{6.9}{0.8}
& \score{8.5}{0.6}
& \second{\score{9.4}{0.8}} \\

\cmidrule(lr){2-3} \cmidrule(lr){4-12}

& \multirow{3}{*}{\textbf{5m\_vs\_6m}}
& Good
& \score{15.8}{3.6}
& \score{2.4}{0.4}
& \score{16.2}{1.6}
& \score{16.8}{2.3}
& \score{16.5}{2.8}
& \score{17.7}{1.1}
& \score{14.7}{2.1}
& \best{\score{18.6}{3.5}}
& \second{\score{18.5}{1.9}} \\
& & Medium
& \score{12.4}{0.9}
& \score{3.8}{0.5}
& \score{15.1}{2.9}
& \score{12.5}{2.1}
& \score{15.2}{2.6}
& \second{\score{16.2}{0.9}}
& \score{12.8}{0.8}
& \score{15.6}{1.3}
& \best{\score{17.9}{1.2}} \\
& & Poor
& \score{7.5}{0.2}
& \score{3.3}{0.5}
& \score{10.5}{3.1}
& \score{8.0}{1.0}
& \score{8.9}{1.3}
& \second{\score{10.8}{0.3}}
& \score{7.7}{0.8}
& \score{9.8}{2.1}
& \best{\score{12.3}{0.9}} \\

\cmidrule(lr){2-3} \cmidrule(lr){4-12}

& \multirow{3}{*}{\textbf{2c\_vs\_64zg}}
& Good
& \score{17.5}{0.4}
& \score{10.1}{0.2}
& \score{12.9}{0.2}
& \score{17.8}{1.3}
& \score{14.7}{2.2}
& \score{16.1}{0.8}
& \score{18.0}{1.3}
& \second{\score{19.1}{0.8}}
& \best{\score{20.2}{0.6}} \\
& & Medium
& \score{12.5}{0.3}
& \score{9.9}{0.2}
& \score{11.6}{0.1}
& \score{10.5}{1.1}
& \score{12.8}{1.2}
& \score{13.9}{0.9}
& \score{11.8}{2.6}
& \second{\score{14.9}{4.1}}
& \best{\score{16.4}{1.0}} \\
& & Poor
& \score{9.7}{0.2}
& \score{9.0}{0.2}
& \score{10.2}{0.1}
& \score{10.2}{2.3}
& \score{10.8}{1.1}
& \best{\score{11.5}{1.1}}
& \score{10.0}{0.3}
& \second{\score{11.4}{0.4}}
& \second{\score{11.4}{1.4}} \\

\cmidrule(lr){2-3} \cmidrule(lr){4-12}

& \multicolumn{2}{c}{\textbf{Average rewards}}
& $12.2$
& $5.5$
& $13.1$
& $13.0$
& $13.8$
& \second{$15.6$}
& $13.4$
& \second{$15.6$}
& \best{$16.7$} \\

\midrule

{\multirow{4}{*}{\STAB{\rotatebox[origin=c]{90}{\textbf{SMACv2}}}}}
& \textbf{terran\_5\_vs\_5}
& replay
& \score{7.3}{1.0}
& \score{13.8}{4.4}
& \score{11.8}{0.9}
& \score{9.3}{0.9}
& \score{13.3}{1.8}
& \second{\score{15.4}{1.3}}
& \score{8.3}{1.9}
& \best{\score{16.6}{4.3}}
& \best{\score{16.6}{4.3}} \\

\cmidrule(lr){2-3} \cmidrule(lr){4-12}

& \textbf{zerg\_5\_vs\_5}
& replay
& \score{6.8}{0.6}
& \second{\score{10.3}{1.2}}
& \second{\score{10.3}{3.4}}
& \score{8.1}{1.7}
& \score{10.2}{1.1}
& \best{\score{12.0}{1.1}}
& \score{4.6}{0.5}
& \score{9.8}{1.5}
& \score{9.7}{2.8} \\

\cmidrule(lr){2-3} \cmidrule(lr){4-12}

& \textbf{terran\_10\_vs\_10}
& replay
& \score{7.4}{0.5}
& \score{12.7}{2.0}
& \score{11.8}{2.0}
& \score{5.5}{1.5}
& \second{\score{13.8}{1.3}}
& \best{\score{14.6}{1.1}}
& \score{5.8}{1.7}
& \score{13.0}{4.7}
& \score{9.4}{2.6} \\

\cmidrule(lr){2-3} \cmidrule(lr){4-12}

& \multicolumn{2}{c}{\textbf{Average rewards}}
& $7.2$
& $12.3$
& $11.3$
& $7.6$
& $12.4$
& \best{$14.0$}
& $6.2$
& \second{$13.1$}
& $11.9$ \\
\bottomrule
\end{tabular}}
\end{table}

\setlength{\fboxsep}{1pt}
\begin{table}[t]
\centering
\vspace{-15pt}
\caption{\textbf{Performance on continuous-action offline MARL benchmarks
(MPE Simple Spread).} \textit{M-R} denotes Medium-Replay.}
\label{tab:mpe_perf}
\scriptsize
\setlength{\tabcolsep}{3.2pt}
\renewcommand{\arraystretch}{1.08}
\resizebox{0.9\textwidth}{!}{
\begin{tabular}{l l c c c c c c c c}
\toprule
\multirow{2}{*}{\textbf{Scenario}}
& \multirow{2}{*}{\textbf{Dataset}}
& \multicolumn{5}{c}{\texttt{Offline MARL}}
& \multicolumn{3}{c}{\texttt{Generative policies}} \\
\cmidrule(lr){3-7} \cmidrule(lr){8-10}
& 
& \texttt{MATD3BC}
& \texttt{MACQL}
& \texttt{ICQ}
& \texttt{OMAR}
& \texttt{OMIGA}
& \texttt{MADiff}
& \texttt{MAC-Flow}
& \texttt{MoSDOT} \\
\midrule

\multirow{4}{*}{\textbf{Spread}}
& Expert
& \second{\score{108.3}{3.3}}
& \score{98.2}{5.2}
& \best{\score{114.9}{2.6}}
& \score{104.0}{3.4}
& \score{80.8}{13.8}
& \score{95.0}{5.3}
& \score{101.7}{10.9}
& \score{104.40}{19.66} \\

& Medium
& \score{29.3}{4.8}
& \score{34.1}{7.2}
& \score{47.9}{18.9}
& \score{29.3}{5.5}
& \score{30.1}{16.9}
& \score{64.9}{7.7}
& \second{\score{80.1}{20.6}}
& \best{\score{95.97}{12.69}} \\

& M-R
& \score{15.4}{5.6}
& \score{20.0}{8.4}
& \score{37.9}{12.3}
& \score{13.6}{5.7}
& \score{5.4}{11.0}
& \score{30.3}{2.5}
& \second{\score{50.4}{33.2}}
& \best{\score{54.12}{15.29}} \\

& Random
& \score{9.8}{4.9}
& \score{24.0}{9.8}
& \second{\score{34.4}{5.3}}
& \score{6.3}{3.5}
& \score{-3.8}{12.3}
& \score{6.9}{3.1}
& \score{31.1}{6.8}
& \best{\score{75.92}{23.81}} \\

\cmidrule(lr){1-2} \cmidrule(lr){3-10}

\multicolumn{2}{c}{\textbf{Average rewards}}
& $40.7$
& $44.1$
& $58.8$
& $38.3$
& $28.1$
& $49.2$
& \second{$65.8$}
& \best{$82.6$} \\

\bottomrule
\end{tabular}}
\vspace{-10pt}
\end{table}

\vspace{-4pt}
\subsection{Offline MARL Benchmark Results}
\label{subsec:benchmarks}
We evaluate MoSDOT on three offline MARL benchmarks.
Tab.~\ref{tab:smac_perf} reports discrete-action results on SMACv1/v2, and Tab.~\ref{tab:mpe_perf} reports continuous-action results on
MPE Simple Spread.
On SMAC, generative policies---particularly flow- and diffusion-based
methods---tend to perform better than Gaussian and behavior cloning (BC) baselines,
consistent with the view that a centralized generative teacher is a
useful starting point. On MPE, MoSDOT shows clearer gains on
\textit{Medium} and \textit{Random}, where replay contains a wider range
of joint actions per observation, and remains comparable to existing
generative baselines on \textit{Expert}, where replay is concentrated on
a single coordinated mode.

These observations are consistent with the view that capacity-matched
source assignment is most useful when replay admits a discernible finite
mode support. The effect is naturally smaller when the data concentrates
on a single coordinated mode (\textit{Expert}), or when the joint-action
space grows faster than per-context replay coverage so that the
identity-style aggregation rule used here approaches sample-level
coupling (e.g., \texttt{terran\_10\_vs\_10}). Designing $q_x$ to recover meaningful mode structure in such regimes
is a natural direction for future work.

\vspace{-5pt}
\section{Conclusion}
\vspace{-5pt}
We studied multi-agent generative distillation through how source noise
is assigned to coordinated joint behaviors during teacher training. Our
analysis shows that endpoint coverage of the teacher is not sufficient
for factorized distillation: routing artifacts at the teacher stage
propagate to decentralized actors through the distillation loss. MoSDOT
addresses this by aligning source noise with a capacity-weighted mode
support via conditional SDOT, separating teacher-side routing artifacts
from the residual product-projection gap that strict-product execution
introduces. A central limitation is the strict-product constraint
itself: independent local sampling cannot represent arbitrary correlated
joint support, and our shared-randomness variant exposes this gap rather
than removing it. Quantifying this residual gap, extending mode-support
routing to longer-horizon and value-reweighted settings, and studying
when shared execution signals are admissible are natural next steps.
\begin{ack}
This work was supported by the Institute of Information \&
Communications Technology Planning \& Evaluation (IITP) grant
funded by the Korea government (MND)
(RS-2026-25553650), the National Research Foundation
of Korea (NRF) grant funded by the Korea government (MSIT)
(No. RS-2023-00208506), and the InnoCORE program of the
Ministry of Science and ICT (N10260099 and N10260110). Prof. Sung-eui Yoon is the corresponding author. 

\paragraph{Competing Interests.}
The authors declare no competing interests.
\end{ack}

\begingroup
\small
\bibliographystyle{unsrtnat}
\bibliography{refs}
\endgroup

\clearpage

\definecolor{mossupport}{RGB}{213,94,0}      %
\definecolor{mosassign}{RGB}{0,114,178}      %
\definecolor{mosteacher}{RGB}{0,158,115}     %
\definecolor{moscritic}{RGB}{204,121,167}    %
\definecolor{mosstudent}{RGB}{86,86,86}      %
\definecolor{mosshared}{RGB}{230,159,0}      %

\providecommand{\mosstage}[2]{%
  \Statex \vspace{0.25em}%
  \Statex \colorbox{#1!12}{\strut\textcolor{#1!70!black}{\textsc{\# #2}}}%
}

\appendix

\section*{Appendix Contents}
\addcontentsline{toc}{section}{Appendix Contents}
\label{app:toc}
\begingroup
\setlength{\parskip}{0.25em}
\setlength{\itemsep}{0pt}
\noindent\textbf{\hyperref[app:proofs]{A\quad Proofs and Mathematical Derivations}}
\begin{itemize}[leftmargin=2.2em,itemsep=0pt,topsep=0pt]
  \item A.1\enspace \hyperref[app:local_projection]{Projection Induced by Local Distillation}
  \item A.2\enspace \hyperref[app:fm_block_locality]{Block-Locality of Ordinary Conditional Flow Matching}
  \item A.3\enspace \hyperref[app:reward_ceiling]{Support Error and Reward Ceiling}
  \item A.4\enspace \hyperref[app:xor_gap]{Strict-Product Gap: A Minimal XOR Example}
  \item A.5\enspace \hyperref[app:shared_randomness_closes_gap]{Shared Randomness Closes the Strict-Product Gap}
\end{itemize}
\noindent\textbf{\hyperref[app:algorithms]{B\quad Algorithms}}
\begin{itemize}[leftmargin=2.2em,itemsep=0pt,topsep=0pt]
  \item B.1\enspace \hyperref[app:alg_mosdot]{MoSDOT (Strict-Product Variant)}
  \item B.2\enspace \hyperref[app:alg_mosdot_sr]{MoSDOT-SR (Shared-Randomness Variant)}
  \item B.3\enspace \hyperref[app:alg_reduction]{Reduction Between the Two Variants}
\end{itemize}
\noindent\textbf{\hyperref[app:limitations]{C\quad Limitations and Discussion}}\par
\noindent\textbf{\hyperref[app:implementation]{D\quad Implementation Details}}
\begin{itemize}[leftmargin=2.2em,itemsep=0pt,topsep=0pt]
  \item D.1\enspace \hyperref[app:aggregation_rule]{Support Aggregation Rule}
  \item D.2\enspace \hyperref[app:finite_cache]{SDOT Solver and Finite-Cache Rebalancing}
  \item D.3\enspace \hyperref[app:teacher_bootstrap]{Teacher Bootstrap for Critic Learning}
  \item D.4\enspace \hyperref[app:architecture]{Network Architecture and Flow Schedule}
\end{itemize}
\noindent\textbf{\hyperref[app:exp_details]{E\quad Experimental Details}}
\begin{itemize}[leftmargin=2.2em,itemsep=0pt,topsep=0pt]
  \item E.1\enspace \hyperref[app:benchmark_details]{Benchmark Environments and Datasets}
  \item E.2\enspace \hyperref[app:baselines]{Baselines}
  \item E.3\enspace \hyperref[app:training_protocol]{Training and Optimization Protocol}
  \item E.4\enspace \hyperref[app:reporting]{Reporting and Statistical Protocol}
  \item E.5\enspace \hyperref[app:compute]{Compute Resources}
  \item E.6\enspace \hyperref[app:metric_defs]{Evaluation Metric Definitions}
\end{itemize}
\noindent\textbf{\hyperref[app:additional_exp]{F\quad Additional Experiments}}
\begin{itemize}[leftmargin=2.2em,itemsep=0pt,topsep=0pt]
  \item F.1\enspace \hyperref[app:student_diagnostics]{Student-Side Landmark Diagnostics}
  \item F.2\enspace \hyperref[app:full_route_locality]{Full Route-Locality Diagnostics}
  \item F.3\enspace \hyperref[app:metric_sensitivity]{Metric Sensitivity Analysis}
\end{itemize}
\endgroup
\clearpage

\section{Proofs and Mathematical Derivations}
\label{app:proofs}

This section collects the formal results that justify the design choices of
MoSDOT. Appendix~\ref{app:local_projection} explains why information that is
hidden from a local actor is averaged out by $\ell_2$ distillation.
Appendix~\ref{app:fm_block_locality} characterizes when conditional flow
matching is block-local at the population optimum, and pinpoints the
assumptions that MoSDOT deliberately breaks.
Appendix~\ref{app:reward_ceiling} converts off-support fan mass into a value
ceiling that even a perfect student cannot escape.
Appendix~\ref{app:xor_gap} shows, on the canonical XOR example, that a clean
teacher alone cannot remove the strict-product gap, motivating the
shared-randomness variant of Section~\ref{sec:method}.
Appendix~\ref{app:shared_randomness_closes_gap} then proves the positive
direction: shared randomness produces a mixture of product distributions that
exactly represents the XOR target, and characterizes the representational
range of this family.

\subsection{Projection Induced by Local Distillation}
\label{app:local_projection}
This section formalizes the claim that the $\ell_2$ distillation loss regresses each local actor onto the conditional mean of the teacher's output given local input. Concretely, we show that the population-optimal local predictor satisfies:
\begin{equation}
\label{eq:local_projection}
f^\star_i(o_i, z_i) \;=\; \mathbb{E}\!\left[\,[\mu_\psi(\textbf{x}, z)]_i \,\big|\, o_i, z_i\,\right],
\end{equation}
so that any component of the teacher's branch decision that remains random after conditioning on $(o_i, z_i)$ is averaged out by the decentralized actor.
For any measurable local predictor $f$, the squared
distillation risk decomposes as:
\begin{align}
\mathbb E\!\left[\|Y-f(S)\|_2^2\right]
&=
\mathbb E\!\left[\left\|Y-\mathbb E[Y\mid S]\right\|_2^2\right]
+\mathbb E\!\left[\left\|\mathbb E[Y\mid S]-f(S)\right\|_2^2\right].
\end{align}
The first term is independent of $f$, and the second term is uniquely
minimized in $L^2$ by:
\[
f^\star(S)=\mathbb E[Y\mid S].
\]
Substituting $Y=g_i(O,z_{1:N})$ and $S=(o_i,z_i)$ gives:
\[
f_i^\star(o_i,z_i)
=
\mathbb E\!\left[g_i(O,z_{1:N})\mid o_i,z_i\right].
\]
Therefore, any component of the teacher branch decision that remains random
after conditioning on the local input is averaged by the decentralized actor.
This is the precise sense in which source assignment affects distillation
compatibility: a teacher branch whose mode label depends on $z_{-i}$ cannot be
recovered by a strict-product local actor that observes only $z_i$.

\subsection{Block-Locality of Ordinary Conditional Flow Matching}
\label{app:fm_block_locality}

This section clarifies when an ordinary conditional flow-matching teacher is
block-local at the population optimum. Let
\[
X_i=(1-T_i)Z_i+T_iA_i,\qquad V_i=A_i-Z_i.
\]
Assume that, conditioned on the joint context $O$, the training law
factorizes across agents:
\[
p(z_{1:N},a_{1:N},t_{1:N}\mid O)
=\prod_{i=1}^{N}p(z_i,a_i,t_i\mid O).
\]
The squared-loss optimal joint flow field is
\[
u_i^\star(O,X_{1:N},T_{1:N})=\mathbb E[V_i\mid O,X_{1:N},T_{1:N}].
\]
Under the conditional factorization above, $(V_i,X_i,T_i)$ is conditionally
independent of the off-agent tuples $(X_{-i},T_{-i})$ given $O$, so
\[
u_i^\star(O,X_{1:N},T_{1:N})=\mathbb E[V_i\mid O,X_i,T_i].
\]
Although a joint network may have access to off-agent chunks, the population
MSE optimum derives no benefit from using them under the factorized training
law. This condition fails precisely when the teacher objective introduces
coupled endpoint assignments, set-level matching, or other gradients that
explicitly depend on full joint branches---which is exactly the regime
MoSDOT operates in.

\subsection{Support Error and Reward Ceiling}
\label{app:reward_ceiling}

This section formalizes the support-preservation claim used in the main text.
For a fixed context $x$, let the target mode-support measure be
\[
\mu_\textbf{x}^\star=\sum_{k=1}^{K}b_k(\textbf{x})\delta_{y_k(\textbf{x})}.
\]
Let $T_{\mathbf x}(z)=y_{\Gamma_{\mathbf x}(z)}(\mathbf x)$ be the
action-valued map induced by the source-to-mode assignment. An exact
assignment satisfies
\[
T_{\mathbf x\#}\rho_0=\sum_{k=1}^{K}\rho_0(\Gamma_\textbf{x}^{-1}(k))\delta_{y_k(\textbf{x})}
=\mu_\textbf{x}^\star.
\]
In the ideal assignment limit the source pushforward has zero mass outside
the prescribed support and matches the target capacities exactly.

Now let $\nu$ be a generated teacher distribution. Let $B_k$ be pairwise
disjoint evaluation regions containing the corresponding support elements
$y_k(\textbf{x})$, and let $F=\mathcal A\setminus\bigcup_{k=1}^{K}B_k$
denote the remaining off-support fan or bridge region. Define
\[
m_k=\nu(B_k),\qquad m_F=\nu(F).
\]
The support error on the quotient partition is
\[
D_{\mathrm{supp}}(\nu,\mu_\textbf{x}^\star)
=\tfrac{1}{2}\!\left(\sum_{k=1}^{K}|m_k-b_k(\textbf{x})|+m_F\right),
\]
which separates capacity error from off-support mass. If a teacher has small
mode-mass error and small fan mass, it is close to the target support
distribution under this quotient evaluation.

Assume a bounded reward or value proxy $R$ such that supported modes are
optimal and the fan region has a reward gap $\Delta>0$:
\[
R(y_k(\textbf{x}))=R^\star,\qquad R(a)\le R^\star-\Delta\quad\forall a\in F.
\]
Then any teacher distribution $\nu$ with fan mass $m_F$ satisfies
\[
J_R(\nu)=\mathbb E_{\nu}[R(a)]\le R^\star-\Delta\,m_F.
\]
If a student $\pi_S$ is within total variation distance $\epsilon$ of $\nu$,
then $\pi_S(F)\ge(m_F-\epsilon)_+$, and therefore
\[
J_R(\pi_S)\le R^\star-\Delta(m_F-\epsilon)_+.
\]
Hence observed off-support mass imposes a value ceiling on faithful
distillation whenever that mass lies in a lower-value region.

\subsection{Strict-Product Gap: A Minimal XOR Example}
\label{app:xor_gap}

This section illustrates why a clean teacher does not imply exact preservation
by a strict-product student. Consider two agents and the target
anti-correlated support
\[
\mathcal M=\{(-1,+1),(+1,-1)\},\qquad
\mu^\star=\tfrac{1}{2}\delta_{(-1,+1)}+\tfrac{1}{2}\delta_{(+1,-1)}.
\]
Let the support reward be $R(x,y)=\mathbf 1[(x,y)\in\mathcal M]$. A strict-product
student matching the marginals of $\mu^\star$ has
\[
q_{\mathrm{clean}}(x,y)=q_x(x)q_y(y),\quad
q_x(\pm 1)=q_y(\pm 1)=\tfrac{1}{2},
\]
so
\[
J_R(q_{\mathrm{clean}})
=q_{\mathrm{clean}}(-1,+1)+q_{\mathrm{clean}}(+1,-1)=\tfrac{1}{2}.
\]
Even a clean non-product teacher cannot be exactly represented by independent
local sampling.

Now consider a fan-contaminated teacher
\[
\nu_m=\tfrac{1-m}{2}\delta_{(-1,+1)}
+\tfrac{1-m}{2}\delta_{(+1,-1)}
+m\,\delta_{(0,0)}.
\]
A strict-product student matching the teacher's marginals independently has
$q_x(\pm 1)=q_y(\pm 1)=(1-m)/2$ and $q_x(0)=q_y(0)=m$, with support reward
\[
J_R(q_m)=\left(\tfrac{1-m}{2}\right)^{\!2}+\left(\tfrac{1-m}{2}\right)^{\!2}
=\tfrac{(1-m)^2}{2}.
\]
Therefore
\[
J_R(q_{\mathrm{clean}})-J_R(q_m)
=\tfrac{1}{2}-\tfrac{(1-m)^2}{2}=m-\tfrac{m^2}{2}>0\quad\text{for }0<m\le 1.
\]
Removing fan mass does not eliminate the intrinsic product gap, but it
strictly improves the marginally distilled product student when fan mass
contaminates the local marginals. This is the analytic counterpart of the
empirical XOR diagnostic in Fig.~\ref{fig:xor_strict_product}; the
complementary positive result---that the shared-randomness extension
(Algorithm~\ref{alg:mosdot_sr}) does close this gap---is given in
Appendix~\ref{app:shared_randomness_closes_gap}.

\subsection{Shared Randomness Closes the Strict-Product Gap}
\label{app:shared_randomness_closes_gap}

Appendix~\ref{app:xor_gap} shows that a strict-product student of the form
Eq.~\eqref{eq:strict_product} cannot represent the anti-correlated
XOR target $\mu^\star=\tfrac{1}{2}\delta_{(-1,+1)}+\tfrac{1}{2}\delta_{(+1,-1)}$.
This section proves the complementary positive direction for the
shared-randomness policy of Eq.~\eqref{eq:shared_randomness_policy}: a
public shared signal $h$ turns the policy into a \emph{mixture of product
distributions}, and this mixture exactly represents anti-correlated supports
such as XOR. We first derive the mixture form from the one-step
implementation, then verify exact XOR reproduction, and finally state the
representational range of the resulting policy family.

\paragraph{From one-step implementation to mixture-of-products.}
Recall the one-step shared-randomness sampler
$\hat a_i=\mu^{\mathrm{sr}}_{w_i}(o_i,h,\epsilon_i)$ with
$h\sim\rho_h$ shared across all agents and $\epsilon_i\sim\rho_i$ drawn
independently per agent. We derive Eq.~\eqref{eq:shared_randomness_policy}
from this implementation in three steps.

\emph{(i) Joint conditional law given all noise.} Since the sampler is
deterministic given $(h,\epsilon_{1:N})$,
\[
p(\mathbf a\mid \textbf{x},h,\epsilon_{1:N})
=\prod_{i=1}^{N}\delta\!\big(a_i-\mu^{\mathrm{sr}}_{w_i}(o_i,h,\epsilon_i)\big).
\]

\emph{(ii) Conditional independence given $h$.} The local noises
$\{\epsilon_i\}$ are mutually independent, so integrating each $\epsilon_i$
against its marginal $\rho_i$ while holding $h$ fixed gives
\[
p(\mathbf a\mid x,h)
=\prod_{i=1}^{N}\underbrace{\int\delta\!\big(a_i-\mu^{\mathrm{sr}}_{w_i}(o_i,h,\epsilon_i)\big)\,\rho_i(d\epsilon_i)}_{=:\,\pi^{\mathrm{sr}}_{w_i}(a_i\mid o_i,h)}
=\prod_{i=1}^{N}\pi^{\mathrm{sr}}_{w_i}(a_i\mid o_i,h).
\]
Conditioning on the shared signal restores per-agent independence: the
local noises were independent to begin with, and $h$ is the only
cross-agent random variable.

\emph{(iii) Marginalization over $h$.} Integrating the shared signal yields
\[
\pi^{\mathrm{sr}}_w(\mathbf a\mid \textbf{x})
=\int p(\mathbf a\mid \textbf{x},h)\,\rho_h(dh)
=\int\prod_{i=1}^{N}\pi^{\mathrm{sr}}_{w_i}(a_i\mid o_i,h)\,\rho_h(dh),
\]
which is exactly Eq.~\eqref{eq:shared_randomness_policy}. The integrand is
a product over agents, but the integral over $h$ produces a
\emph{mixture} of such products. Conditional independence given $h$ does
not imply marginal independence, so the joint law generally satisfies
$a_i\not\!\perp a_j$ even though $a_i\perp a_j\mid h$.

\paragraph{Exact XOR reproduction.}
For the target
$\mu^\star=\tfrac{1}{2}\delta_{(-1,+1)}+\tfrac{1}{2}\delta_{(+1,-1)}$
of Appendix~\ref{app:xor_gap}, take a binary shared signal
$h\in\{0,1\}$ with $\rho_h(0)=\rho_h(1)=\tfrac{1}{2}$ and degenerate local
noises ($d_{\epsilon_i}=0$). Define the per-agent maps
\[
\mu^{\mathrm{sr}}_{w_1}(\cdot,h,\cdot)=\begin{cases}-1 & h=0,\\+1 & h=1,\end{cases}
\qquad
\mu^{\mathrm{sr}}_{w_2}(\cdot,h,\cdot)=\begin{cases}+1 & h=0,\\-1 & h=1.\end{cases}
\]
Substituting into Eq.~\eqref{eq:shared_randomness_policy} gives
\[
\pi^{\mathrm{sr}}_w
=\tfrac{1}{2}\,\delta_{-1}\otimes\delta_{+1}
+\tfrac{1}{2}\,\delta_{+1}\otimes\delta_{-1}
=\mu^\star.
\]
Each agent acts deterministically from its own input alone, yet the joint
law is exactly the anti-correlated target. The only cross-agent channel is
the public draw of $h$; no observations, actions, or messages are exchanged.

\paragraph{Representational range.}
The construction generalizes to any target that factorizes through a
discrete latent $K$,
\[
\mu^\star(\mathbf a\mid \textbf{x})=\sum_{k=1}^{K}\pi_k(\textbf{x})\,\prod_{i=1}^{N}\mu^\star_{i,k}(a_i\mid o_i,\textbf{x}),
\]
by taking $\rho_h$ supported on $\{1,\ldots,K\}$ with $\rho_h(k)=\pi_k(\textbf{x})$
and $\pi^{\mathrm{sr}}_{w_i}(\cdot\mid o_i,k)=\mu^\star_{i,k}(\cdot\mid o_i,\textbf{x})$.
A shared dimension of $\lceil\log_2 K\rceil$ bits suffices for any
$K$-component mixture-of-products target, and richer $\rho_h$ (e.g.\
$\rho_h=\rho_0$ on $\mathbb R^d$) yields a universal class of correlated
decentralized policies under independent local computation and a single
public coordination channel.

\paragraph{Connection to the MoSDOT-SR algorithm.}
Algorithm~\ref{alg:mosdot_sr} uses the shared signal $h$ as the public latent in
this mixture-of-products view. In the Brenier-polar diagnostic instantiation,
$h$ is chosen so that the cached SDOT assignment $\Gamma_{\mathbf x}(h)$ selects
a replay-derived support element, with mixture weights induced by the capacities
$b_k(\mathbf x)$. The generic algorithm does not require exposing the entire
teacher source or committing to a particular cell-lookup parameterization; it
only requires that all agents receive the same public signal $h$ while retaining
local computation.

\par\noindent\begin{minipage}{\linewidth}
\section{Algorithms}
\label{app:algorithms}

This section gives compact pseudocode for the two MoSDOT variants. The mode
support, capacities, and SDOT source assignment are already defined in
Sec.~\ref{subsec:mode_support}--\ref{subsec:teacher_source_assignment}; the
algorithms therefore call these steps as subroutines rather than expanding the
SDOT dual solver or finite-cache implementation. Aggregation choices, cache
rebalancing, and teacher-bootstrap details are deferred to
Appendix~\ref{app:implementation}.

\subsection{MoSDOT (Strict-Product Variant)}
\label{app:alg_mosdot}

\begin{algorithm}[H]
\caption{MoSDOT: SDOT-aligned teacher training and strict-product distillation.}
\label{alg:mosdot}
\small
\begin{algorithmic}[1]
\Require offline replay dataset $\mathcal D$; source law $\rho_0$ over
$\mathbf z=(z_1,\ldots,z_N)$; support aggregation rule $q_{\mathbf x}$;
cost $c_{\mathbf x}$; teacher field $u_\psi$; decentralized actors
$\{\mu_{w_i}\}_{i=1}^{N}$; centralized critic $Q_{\rm tot}$ and target
$\bar Q_{\rm tot}$.
\Ensure decentralized actors $\{\mu_{w_i}\}_{i=1}^{N}$.

\mosstage{mossupport}{Build routed mode support}
\For{each cached conditioning context $\mathbf x$}
    \State $(\mathcal Y_{\mathbf x}, b(\mathbf x))
    \leftarrow \textsc{AggregateSupport}(\mathcal D,\mathbf x; q_{\mathbf x})$
    \Comment{Eqs.~\eqref{eq:target_joint_action_set}--\eqref{eq:mode_support_set}}
    \State $\Gamma_{\mathbf x}\leftarrow
    \textsc{SolveSDOT}(\rho_0,\mathcal Y_{\mathbf x},b(\mathbf x);c_{\mathbf x})$
    \Comment{Eqs.~\eqref{eq:mosdot_primal}--\eqref{eq:source_to_mode_assignment}}
\EndFor

\While{not converged}
    \State Sample replay batch
    $\mathcal B=\{(\mathbf x_b,\mathbf o_b,\mathbf a_b,r_b,\mathbf x'_b,\mathbf o'_b,d_b)\}_{b=1}^{B}\sim\mathcal D$.

    \mosstage{mosteacher}{Train SDOT-aligned teacher}
    \State Sample $\mathbf z_b\sim\rho_0$ and $t_b\sim\mathrm{Unif}([0,1])$.
    \State Set routed target $\mathbf y_b\leftarrow
    \mathbf y_{\Gamma}(\mathbf x_b,\mathbf z_b)$,
    \Statex \hspace{\algorithmicindent}interpolation
    $\bar{\mathbf a}_{b,t}\leftarrow(1-t_b)\mathbf z_b+t_b\mathbf y_b$, and
    \Statex \hspace{\algorithmicindent}displacement
    $\mathbf v_b\leftarrow \mathbf y_b-\mathbf z_b$.
    \State Update $u_\psi$ using the teacher loss in
    Eq.~\eqref{eq:assignment_teacher_loss}.
    \State Define the teacher sampler
    $\mu_\psi(\mathbf x,\mathbf z)\leftarrow\mathrm{Flow}_{u_\psi}(\mathbf x,\mathbf z)$.

    \mosstage{moscritic}{Train centralized critic}
    \State Sample $\mathbf z'_b\sim\rho_0$ and bootstrap
    $\mathbf a'_b\leftarrow\mu_\psi(\mathbf x'_b,\mathbf z'_b)$.
    \State Update $Q_{\rm tot}$ with the teacher-bootstrap Bellman loss
    (Appendix~\ref{app:teacher_bootstrap}) and soft-update $\bar Q_{\rm tot}$.

    \mosstage{mosstudent}{Distill strict-product actors}
    \State Sample fresh joint source $\boldsymbol\zeta_b=(\zeta_{b,1},\ldots,\zeta_{b,N})\sim\rho_0$.
    \State Compute
    $\tilde{\mathbf a}_b\leftarrow\mu_\psi(\mathbf x_b,\boldsymbol\zeta_b)$ and
    $\hat{\mathbf a}_b\leftarrow
    (\mu_{w_i}(o_{b,i},\zeta_{b,i}))_{i=1}^{N}$.
    \State Update $\{w_i\}_{i=1}^{N}$ using the value-guided distillation
    objective in Eq.~\eqref{eq:value_guided_actor_loss}.
\EndWhile

\State \Return $\{\mu_{w_i}\}_{i=1}^{N}$.
\end{algorithmic}
\end{algorithm}
\end{minipage}\par

\noindent\begin{minipage}{\linewidth}
\subsection{MoSDOT-SR (Shared-Randomness Variant)}
\label{app:alg_mosdot_sr}

Algorithm~\ref{alg:mosdot_sr} keeps the same mode-support, SDOT, teacher, and
critic stages as Algorithm~\ref{alg:mosdot}. It only changes the actor
signature by splitting the teacher source into a public component $h$ and
private local noises $\epsilon_i$, matching the shared-randomness policy in
Eq.~\eqref{eq:shared_randomness_policy}.

\begin{algorithm}[H]
\caption{MoSDOT-SR: shared-randomness distillation and execution.}
\label{alg:mosdot_sr}
\small
\begin{algorithmic}[1]
\Require trained MoSDOT teacher $\mu_\psi$ and critic $Q_{\rm tot}$; source
factorization $\rho_0=\rho_h\otimes\bigotimes_{i=1}^{N}\rho_i$; decentralized
actors $\{\mu^{\mathrm{sr}}_{w_i}(o_i,h,\epsilon_i)\}_{i=1}^{N}$; public seed
generator $G$.
\Ensure decentralized actors $\{\mu^{\mathrm{sr}}_{w_i}\}_{i=1}^{N}$.

\mosstage{mosstudent}{Distill shared-randomness actors}
\While{not converged}
    \State Sample replay batch $\mathcal B\sim\mathcal D$, shared signals
    $h_b\sim\rho_h$, and local noises $\epsilon_{b,i}\sim\rho_i$.
    \State Set $\mathbf z_b=(h_b,\epsilon_{b,1:N})$ and compute
    $\tilde{\mathbf a}_b\leftarrow\mu_\psi(\mathbf x_b,\mathbf z_b)$.
    \State Compute student action
    $\hat{\mathbf a}_b\leftarrow
    (\mu^{\mathrm{sr}}_{w_i}(o_{b,i},h_b,\epsilon_{b,i}))_{i=1}^{N}$.
    \State Update $\{w_i\}_{i=1}^{N}$ using the same value-guided
    distillation objective as Algorithm~\ref{alg:mosdot}.
\EndWhile

\mosstage{mosshared}{Decentralized execution}
\For{execution step $n$}
    \State All agents publicly reconstruct $h_n=G(s_n)\sim\rho_h$; each
    agent $i$ privately draws $\epsilon_{i,n}\sim\rho_i$.
    \State Agent $i$ outputs
    $a_{i,n}=\mu^{\mathrm{sr}}_{w_i}(o_{i,n},h_n,\epsilon_{i,n})$.
\EndFor

\State \Return $\{\mu^{\mathrm{sr}}_{w_i}\}_{i=1}^{N}$.
\end{algorithmic}
\end{algorithm}
\end{minipage}\par

\subsection{Reduction Between the Two Variants}
\label{app:alg_reduction}

Algorithm~\ref{alg:mosdot} is recovered from Algorithm~\ref{alg:mosdot_sr}
when no public randomness is used. Equivalently, either the shared component is
degenerate ($d_h=0$) so that $\rho_0=\bigotimes_i\rho_i$, or the actor ignores
$h$, i.e., $\mu^{\mathrm{sr}}_{w_i}(o_i,h,\epsilon_i)=\mu_{w_i}(o_i,\epsilon_i)$.
Then Eq.~\eqref{eq:shared_randomness_policy} reduces to the strict-product
policy in Eq.~\eqref{eq:strict_product}. At the other endpoint, exposing a
non-degenerate public signal makes the deployed law a mixture of product
policies, which is the mechanism analyzed in Appendix~\ref{app:shared_randomness_closes_gap}.

\section{Limitations and Discussion}
\label{app:limitations}

We close by collecting the main limitations of MoSDOT and noting partial
remedies where they exist.

\paragraph{Strict-product execution is fundamentally limited.}
Independent local sampling cannot represent arbitrary correlated joint
support: even with a perfect teacher, the strict-product student of
Eq.~\eqref{eq:strict_product} fails on anti-correlated targets such as
XOR (Appendix~\ref{app:xor_gap}). MoSDOT improves the centralized
teacher but does not remove this execution-time constraint. The
shared-randomness variant (Appendix~\ref{app:shared_randomness_closes_gap})
closes the gap on supports representable as mixtures of products, but
requires a public coordination channel, which is not always admissible
in deployment. Quantifying when such a channel is available---and how
much shared dimension is actually needed---remains open.

\paragraph{Empirical scope.} We evaluate on three offline MARL benchmarks (SMACv1, SMACv2, MPE Simple Spread) with at most $10$ agents and either small discrete or low-dimensional continuous action spaces. Scaling to larger agent populations, richer action spaces, or online fine-tuning is left for future work, as is direct comparison with communication-augmented baselines that use richer cross-agent channels than the public seed considered here.

\section{Implementation Details}
\label{app:implementation}

This section documents implementation choices that are specific to MoSDOT
itself, as opposed to general experimental settings that apply to all
methods (those are deferred to Appendix~\ref{app:exp_details}).

\subsection{Support Aggregation Rule}
\label{app:aggregation_rule}

The aggregation rule $q_\textbf{x}$ specifies the granularity of the replay-derived
target support and is not treated as an oracle mode label. The identity rule
$q_\textbf{x}(m)=m$ is always admissible and recovers the raw sample support. When
repeated, quantized, or high-confidence clustered joint targets are present,
$q_\textbf{x}$ may merge those targets into a smaller support. We use \emph{conservative
aggregation}: ambiguous targets remain separate rather than being forced into
a shared support element. The representation used by $q_\textbf{x}$ can be a joint
action, an action chunk, a short-horizon endpoint, or a learned target
embedding. In all cases the support index refers to a full joint target rather
than independently chosen per-agent modes. This convention prevents the
assignment problem from degenerating into a product of local mode assignments.
For discrete-action benchmarks we use the identity rule; for continuous-action
benchmarks we use a simple joint-action quantization at fixed bin width.

\subsection{SDOT Solver and Finite-Cache Rebalancing}
\label{app:finite_cache}

The Laguerre assignment $\Gamma_x$ defined in Sec.~\ref{sec:method} is
specified for the population source law $\rho_0$. In implementation we
approximate $\rho_0$ by a finite Monte-Carlo cache: for each
context/support group $c$ we draw
$\mathcal Z_c=\{z^\ell\}_{\ell=1}^{L}\stackrel{\text{i.i.d.}}{\sim}\rho_0$
of size $L$ ($65{,}536$ in the landmark diagnostic) and define the empirical cell masses
\[
\hat m_{c,k}^{(L)}
=\tfrac{1}{L}\,\big|\{\ell:\Gamma_c(z^\ell)=k\}\big|.
\]
The cost $c_c$ is the squared Euclidean distance in the joint-action
representation that defines $\mathcal Y_c$.

\paragraph{Dual solver.}
The SDOT dual is a $K_c$-dimensional concave program. We estimate the dual
weights using stochastic gradient updates with a fixed iteration budget,
then evaluate the resulting assignment on a separately sampled finite
Gaussian cache.

\paragraph{Finite-cache rebalancing.}
With finite $L$, residual deviations between $\hat m_{c,k}^{(L)}$ and
$b_{c,k}$ may persist after dual optimization. Following the
rebalancing principle of~\citet{kong2025alignflowimproving}, we preserve
as many original source--target assignments as possible while adjusting
target counts to match the prescribed frequencies up to integer rounding.
For uniform target weights, the rebalanced counts differ by at most one.
This step
acts only on the finite cache and does not modify the population SDOT
problem; in our landmark caches it reassigns $0.6\%$ to $1.6\%$ of
cached entries.

\subsection{Teacher Bootstrap for Critic Learning}
\label{app:teacher_bootstrap}

In our value-guided implementation, the assignment-trained teacher provides
bootstrap actions for critic learning. For a replay transition
$(o_i,a_i,r_i,o_i')$ we sample a joint next action from the teacher,
\[
\mathbf a'=\mu_\psi(o',z'),\qquad z'\sim\rho_0,
\]
and train an individual critic with
\[
\mathcal L_Q(\theta_i)
=\mathbb E\!\left[\Big(Q_{\theta_i}(o_i,a_i)
-\big(r_i+\gamma(1-d)\bar Q_{\bar\theta_i}(o_i',a_i')\big)\Big)^{\!2}\right].
\]
The decentralized actors are then updated with the value-guided distillation
objective in Eq.~\eqref{eq:value_guided_actor_loss}. This teacher-bootstrap variant
is an implementation choice; the core MoSDOT construction is the source
assignment used to train the centralized teacher.

\subsection{Network Architecture and Flow Schedule}
\label{app:architecture}

All networks are $[512,512,512,512]$ MLPs with layer normalization. The
teacher conditions on the joint observation and joint noise; the
decentralized students take only their local observation and local noise
slice. The MoSDOT-SR variant additionally uses a trainable noise map
$r_{\eta_i}(o_i,h,\epsilon_i)$ that combines the public shared signal $h$
with the local noise $\epsilon_i$. The teacher uses a linear-interpolation
flow with $10$ Euler steps for both training-time bootstrap actions and
inference.

\section{Experimental Details}
\label{app:exp_details}

\subsection{Benchmark Environments and Datasets}
\label{app:benchmark_details}

We evaluate on three offline MARL benchmarks. The choice of maps, quality
regimes, and dataset sources follows the conventions of \citet{lee2026macflow}
and \citet{formanek2023off} so that comparator scores remain directly
comparable.

\paragraph{SMACv1.}
SMACv1 \citep{samvelyan2019starcraft} simulates decentralized StarCraft~II
combat between a built-in AI and learned policies. Agents observe partial
local information within a sight range, and the discrete action space includes
movement, target attack, stop, and no-op. We evaluate on five maps spanning
homogeneous and heterogeneous unit compositions: 3m, 8m, 2s3z, 5m\_vs\_6m, and
2c\_vs\_64zg. For each map we use the Good, Medium, and Poor quality datasets
from the off-the-grid benchmark \citep{formanek2023off}, giving
$15$~dataset--scenario pairs.

\paragraph{SMACv2.}
SMACv2 \citep{ellis2023smacv2} extends SMACv1 with randomized start positions,
randomized unit types, and rebalanced sight and attack ranges, increasing the
diversity of episodes. We evaluate on three maps under the Replay regime:
terran\_5\_vs\_5, zerg\_5\_vs\_5, and terran\_10\_vs\_10. Datasets are again
drawn from the off-the-grid benchmark.

\paragraph{MPE Simple Spread.}
MPE \citep{lowe2017multi} is a continuous-action particle environment for
cooperative coordination. In Simple Spread, agents must cover landmarks while
avoiding collisions. We use the four quality regimes (Expert, Medium,
Medium-Replay, Random) collected by \citet{pan2022plan}. Following
\citet{lee2026macflow}, returns are normalized as
\[
100\times\frac{S-S_{\text{random}}}{S_{\text{expert}}-S_{\text{random}}},
\]
with $S_{\text{random}}=159.8$ and $S_{\text{expert}}=516.8$.

\subsection{Baselines}
\label{app:baselines}

\paragraph{Diagnostic baselines.}
On the landmark and XOR diagnostics, we compare against three teacher
parameterizations. \textit{Independent flow matching} \citep{lipman2022flow}
pairs source noise and replay targets independently within each context.
\textit{IMLE policy} \citep{rana2025imlepolicy} uses a set-level objective based
on implicit maximum likelihood estimation to improve endpoint coverage and
reduce mode collapse. \textit{Drifting} \citep{deng2026drifting} trains
generated samples through attraction toward the target support and
repulsion among generated samples.

\paragraph{Benchmark baselines.}
For benchmark evaluation we additionally compare against three categories of
offline MARL baselines reported in \citet{lee2026macflow}, all reproduced
under the same dataset splits and seed protocol.

\emph{Gaussian-policy baselines.} BC is a behavioral cloning policy
parameterized as a Gaussian (continuous) or categorical (discrete) head.
MABCQ \citep{jiang2021offline} extends BCQ \citep{fujimoto2019off} to
multi-agent settings with a Q-mixer. MACQL extends CQL
\citep{kumar2020conservative} with a conservative penalty. MATD3BC extends
TD3+BC \citep{fujimoto2021minimalist} with $\alpha=2.5$. ICQ
\citep{yang2021believe} uses an implicit constraint approach for joint policy
regularization. OMAR \citep{pan2022plan} combines CQL-style critic
regularization with a CEM-based policy improvement head. OMIGA
\citep{wang2023offline} uses a state-dependent value mixer with
advantage-weighted regression \citep{peng2019advantage}.

\emph{Diffusion-policy baselines.} Diffusion BC is a behavioral cloning
policy with a $200$-step Gaussian DDPM scheduler. MADiff
\citep{zhu2024madiff} models coordinated multi-agent trajectories with a
centralized-then-decentralized diffusion policy. DoF \citep{li2025dof}
factorizes the diffusion process across agents under the IGM-compatible
W-concat factorization with $200$ denoising steps.

\emph{Flow-policy baselines.} Flow BC trains an independent flow policy
per agent with the same $10$-step linear-interpolation schedule as
MAC-Flow but without the joint teacher and IGM critic. MAC-Flow
\citep{lee2026macflow} is the immediate predecessor of MoSDOT and shares
the centralized-teacher to decentralized one-step actor pipeline. MoSDOT
replaces the independent source--target pairing of MAC-Flow's teacher
with the SDOT-aligned coupling.

All baselines use the same $[512,512,512,512]$ MLP architecture as MoSDOT
and follow the official implementations cited above. Public repositories
used for reimplementation are listed in the supplementary code release.

\subsection{Training and Optimization Protocol}
\label{app:training_protocol}

\paragraph{Optimization.}
We use Adam with learning rate $3\times 10^{-4}$ for both actor and
critic, batch size $64$, weight decay $0$, optimizer epsilon $10^{-5}$,
Polyak averaging coefficient $\tau=0.005$, and discount factor
$\gamma=0.995$. The BC distillation coefficient is $\alpha=3.0$ by default.

\paragraph{Training schedule.}
We train for $10^6$ gradient steps on SMACv1 and SMACv2, and
$5\times 10^5$ steps on MPE, matching the budget of \citet{lee2026macflow}.
We evaluate every $50$k steps with $10$ evaluation episodes.

\subsection{Reporting and Statistical Protocol}
\label{app:reporting}

For benchmark scores we run $6$ random seeds and report mean $\pm$ two
standard deviations following \citet{lee2026macflow}. Unless otherwise
specified, diagnostic $\pm$ values denote one standard deviation across
$4$ seeds. Tables~\ref{tab:full_route_locality}
and~\ref{tab:metric_sensitivity_perseed} use seed--agent aggregation for
the metrics identified in their captions. Best and second-best values are
highlighted according to their numerical ranking, where indicated.

In Fig.~\ref{fig:route_locality} (left), bars show mean joint mode coverage
and error bars indicate one standard deviation across four seeds; hatched
and solid bars denote teachers and students, respectively. The right panel
shows teacher Jacobian ratios, with one agent-averaged value per seed.
Boxes span the interquartile range, central lines mark medians, and whiskers
extend to the most extreme observations within $1.5$ interquartile ranges
of the box; points show the individual seed values.

\subsection{Compute Resources}
\label{app:compute}

All experiments run on a single NVIDIA RTX 4090 GPU with CUDA~12.2 and
JAX~0.4.30. SDOT assignments are precomputed before teacher training and
reused throughout optimization.

\paragraph{Preprocessing and training time.}
Table~\ref{tab:landmark_wallclock} compares standard flow matching and
MoSDOT on the landmark diagnostic under matched data, architecture,
minibatch size, and training budget. Both methods require dataset
preparation; MoSDOT additionally performs dual optimization, Laguerre
assignment, finite-cache rebalancing, and cache storage. The additional
preprocessing is paid once, and the deployed actor remains the same
one-step network.

\begin{table}[H]
\centering
\footnotesize
\captionsetup{position=bottom,skip=6pt}
\setlength{\tabcolsep}{7pt}
\renewcommand{\arraystretch}{1.2}
\begin{tabularx}{\linewidth}{@{}Xrrr@{}}
\toprule
Method & \makecell[r]{Preprocessing\\(s)}
       & \makecell[r]{Training\\(min)} & \makecell[r]{Total\\(min)} \\
\midrule
Standard flow matching & 70 & $37.75\pm0.01$ & 38.9 \\
\textbf{MoSDOT} & 279 & $38.88\pm0.29$ & 43.5 \\
\bottomrule
\end{tabularx}
\caption{\textbf{Landmark wall-clock comparison.} Training uses $40{,}000$
steps; training times are mean $\pm$ one standard deviation over four seeds.
Total time includes one preprocessing pass and the mean training time.}
\label{tab:landmark_wallclock}
\end{table}

\paragraph{Occupied support and cache construction.}
Table~\ref{tab:support_cache_cost} reports retained replay rows $M$,
occupied support elements $K$, and one-time cache-construction times for
the measured benchmark configurations. Support size is bounded by the
retained replay data ($K\leq M$), rather than the full Cartesian
joint-action space. It need not increase monotonically with the number
of agents: replay diversity and aggregation granularity also affect $K$.

\begin{table}[H]
\centering
\footnotesize
\captionsetup{position=bottom,skip=6pt}
\setlength{\tabcolsep}{4pt}
\renewcommand{\arraystretch}{1.2}
\begin{tabularx}{\linewidth}{@{}c>{\raggedright\arraybackslash}Xrrrr@{}}
\toprule
Agents & Dataset & \makecell[r]{Retained\\rows $M$}
       & \makecell[r]{Occupied\\support $K$} & $K/M$
       & \makecell[r]{Cache\\time (s)} \\
\midrule
3 & SMACv1 3m / Poor & 200,000 & 6,625 & 3.3\% & 10.5 \\
3 & MPE Spread / Medium & 5,000,000 & 72,509 & 1.5\% & 122.4 \\
5 & SMACv1 5m\_vs\_6m / Good & 200,000 & 20,454 & 10.2\% & 10.6 \\
5 & SMACv2 zerg\_5\_vs\_5 / Replay & 200,000 & 2,380 & 1.2\% & 38.6 \\
8 & SMACv1 8m / Poor & 200,000 & 32,653 & 16.3\% & 67.1 \\
10 & SMACv2 terran\_10\_vs\_10 / Replay & 200,000 & 88,729 & 44.4\% & 147.7 \\
\bottomrule
\end{tabularx}
\caption{\textbf{Occupied replay support and one-time cache cost.}
$M$ and $K$ are totals over the retained dataset, not counts of all
possible joint actions. Times measure cache construction before training.}
\label{tab:support_cache_cost}
\end{table}

\paragraph{Assignment cost versus the complete dual solve.}
For an SDOT problem with $K$ occupied support elements in a $d$-dimensional
joint-action representation, assigning $L$ source samples by exhaustive
cost comparison takes $O(LKd)$ operations. With $T$ stochastic dual updates
using $B$ source samples per update, dual optimization requires
$O(TBKd)$ distance-evaluation work; assigning the final cache adds
$O(LKd)$. Thus the dominant distance-evaluation cost is
\[
O\!\left((TB+L)Kd\right).
\]
For multiple conditioning groups, these costs add across groups. The
measured cache times also include preprocessing, rebalancing, and storage.
With fixed per-agent action dimension, $d$ grows linearly with the number
of agents, while $K$ depends on the occupied replay support. Neither the
dual solve nor cache construction is repeated during decentralized execution.

\subsection{Evaluation Metric Definitions}
\label{app:metric_defs}

This subsection defines the diagnostic metrics used in
Section~\ref{sec:experiments} and Appendix~\ref{app:additional_exp}.
Teacher-side metrics (Appendix~\ref{app:teacher_metrics}) test whether
the teacher reaches the mode support, and route-locality metrics
(Appendix~\ref{app:route_locality_metrics}) test whether mode choices
depend only on locally observable information.

\subsubsection{Teacher-Side Support Metrics}
\label{app:teacher_metrics}

\paragraph{Fan-free.}
For the landmark diagnostic, an agent's initial action is valid if its
direction lies within $10^\circ$ of either of its two landmark directions;
actions with norm below $10^{-6}$ are invalid. Valid actions receive the
nearest-direction branch label. \textit{Fan-free} is the valid-action
fraction, averaged over agents. Higher is better.

\paragraph{Balance.}
Let $n_{i,0}$ and $n_{i,1}$ count valid actions assigned to agent $i$'s
two local modes. \textit{Balance} reports the minimum across agents of
\[
\frac{2\min(n_{i,0},n_{i,1})}{n_{i,0}+n_{i,1}},
\]
with value $0$ if no action is valid. Higher is better.

\paragraph{Success.}
\textit{Success} is the fraction of agent trajectories that come within
distance $0.05$ of either of that agent's landmarks at any time during
the five-step rollout, averaged over agents and rollouts.

\paragraph{Route consistency.}
Route consistency is the product of the fan-free fraction and one minus
the agent-averaged KNN flip rate at $k=20$, defined below. Thus it rewards
both valid actions and stable local branch labels.

\paragraph{Joint mode coverage.}
Let $V$ denote the event that all agents' actions are valid, and let
$\mathbf m=(m_1,\ldots,m_N)$ be their joint branch-label tuple. Joint
mode coverage is $\Pr(V)\,H(\mathbf m\mid V)/\log_2(2^N)$, with entropy
measured in bits and value $0$ when no joint sample is valid.

\subsubsection{Route-Locality Metrics}
\label{app:route_locality_metrics}

Route locality is reported through two complementary families. The
sample-level family operates on replayed teacher samples. The model-level
family operates on the restored teacher MLP and provides causal evidence
for cross-noise dependence.

\paragraph{Own-cell entropy.}
For each agent $i$, samples with valid branch labels $m_i\in\{0,1\}$
are retained, and their $z_i$ values are partitioned into a $b\times b$
quantile grid with default $b=10$. We report the cell-mass-weighted
average of empirical branch entropy in bits, normalized over retained
cells. Cells require at least $20$ samples in
Table~\ref{tab:full_route_locality} and $4$ in the sensitivity analysis
of Table~\ref{tab:metric_sensitivity_perseed}. The own-cell mix rate uses
the same cells and weights, replacing entropy by $\min(p,1-p)$, where
$p$ is the fraction of branch-$1$ labels within a cell.

\paragraph{KNN flip rate.}
Among samples with valid branch labels, we find each sample's $k$
nearest neighbors in $z_i$, excluding itself, with default $k=20$.
We report the mean fraction of neighbors with a different branch
label $m_i$. Sensitivity at $k\in\{10,20,50\}$ is reported in
Appendix~\ref{app:metric_sensitivity}.

\paragraph{Off-diagonal AUC excess.}
For each $j\ne i$, we fit a Fisher linear discriminant to predict valid
branch labels $m_i$ from $z_j$ and evaluate its AUC on those samples.
We report $\max_{j\ne i}\max(\mathrm{AUC}_{ij},1-\mathrm{AUC}_{ij})-0.5$.
A value near zero indicates little linearly decodable cross-noise
dependence.

\paragraph{Jacobian off-diagonal ratio.}
On the restored teacher, we differentiate the full endpoint sampler
with respect to source noise via forward-mode autodiff. Writing
$J_{ij}^{(s)}=\partial a_i^{(s)}/\partial z_j^{(s)}$ for sample $s$,
the reported ratio is
\[
\frac{\mathbb{E}_{s,i,j\ne i}\|J_{ij}^{(s)}\|_F}
{\mathbb{E}_{s,i}\|J_{ii}^{(s)}\|_F},
\]
where each expectation is a uniform empirical average, with a numerical
floor on the denominator. The ratio uses the
unclipped endpoint helper so that action saturation does not mask
cross-noise sensitivity. The student ratio is exactly $0$ by architecture
and serves as a structural reference.

\paragraph{Counterfactual flip rate.}
We fix $z_i$, resample $z_{-i}$ from $\rho_0$ $32$ times per sample, and
pass each through the restored teacher sampler. Among pairs with valid
branch labels for both the original and resampled output, the flip rate
is the fraction that change agent $i$'s branch label. This is the causal
counterpart of the KNN flip diagnostic. It tests
whether the model itself uses $z_{-i}$ to select agent $i$'s branch,
rather than whether the population of trained samples happens to mix.

\section{Additional Experiments}
\label{app:additional_exp}

This section collects supplementary numerical results referenced from the
main text. Appendices~\ref{app:student_diagnostics}--\ref{app:metric_sensitivity}
report extended diagnostic results on the controlled tasks: student-side
endpoint quality, full sample- and model-level route-locality (entanglement)
diagnostics, and a sensitivity study showing that the route-locality ranking
is invariant to bin and neighborhood choices.

\subsection{Student-Side Landmark Diagnostics}
\label{app:student_diagnostics}

Tab.~\ref{tab:landmark_endpoint_student} reports endpoint-quality metrics
for one-step factorized students distilled from each teacher under the same
student architecture, seed list, and training budget. Comparing this table
with the corresponding teacher-side numbers in Tab.~\ref{tab:landmark_endpoint}
quantifies the teacher-to-student transfer described in
Section~\ref{sec:experiments}: the MoSDOT student largely retains the
teacher's fan-free, success, and balance scores, whereas IMLE and Drifting students lose
substantial endpoint quality through distillation. This is consistent with
the projection result of Appendix~\ref{app:local_projection}---only teachers
whose mode choice is locally identifiable can be reproduced by a strict-product
student. Figure~\ref{fig:joint_mode_tuple_support} shows the corresponding
joint mode-tuple distributions.

\par\noindent\begin{minipage}{\linewidth}
\begin{figure}[H]
  \centering
  \includegraphics[width=\linewidth]{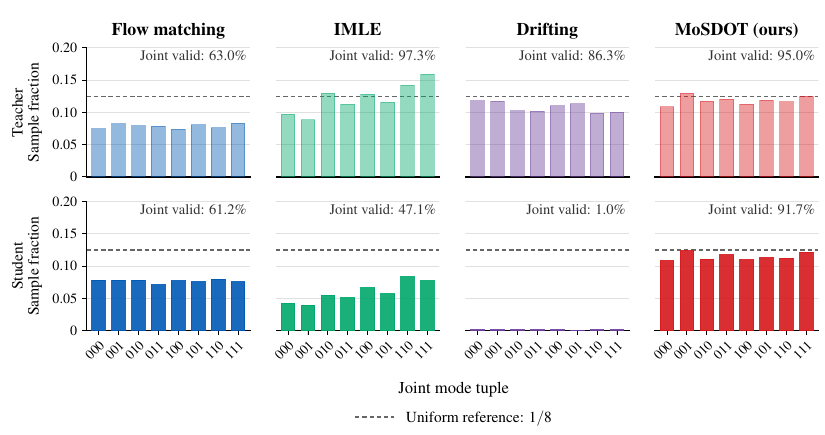}
  \caption{\textbf{Joint mode-tuple support, teacher vs.\ student.}
  Bars show the fraction of $32{,}768$ samples (fan included in the
  denominator) falling into each of the 8 joint mode tuples on the
  3-agent landmark task; the dashed line marks the zero-fan uniform
  ideal $1/8$. Top row: centralized teacher (joint flow ODE). Bottom row:
  one-step factorized student distilled from each teacher under matched
  architecture and budget. MoSDOT preserves both validity and balance
  through distillation (95.0\%\,$\to$\,91.7\% valid), while IMLE loses
  half its valid mass (97.3\%\,$\to$\,47.1\%) and Drift collapses
  entirely to the fan region (86.3\%\,$\to$\,1.0\%). This is the
  per-mode visualization of the aggregate scores in
  Tab.~\ref{tab:landmark_endpoint_student} and the empirical counterpart
  of the projection result in Appendix~\ref{app:local_projection}.}
  \label{fig:joint_mode_tuple_support}
\end{figure}

\begin{table}[H]
\centering
\footnotesize
\captionsetup{position=bottom,skip=6pt}
\setlength{\tabcolsep}{6pt}
\renewcommand{\arraystretch}{1.2}
\newcommand{\studentscore}[2]{\mbox{#1\,{\scriptsize$\pm#2$}}}
\newcommand{\studentbest}[2]{\cellcolor{tblBest}\studentscore{\textbf{#1}}{#2}}
\newcommand{\studentsecond}[2]{\cellcolor{tblSecond}\studentscore{\underline{#1}}{#2}}
\begin{tabular}{@{}lccc@{}}
\toprule
Method & Fan-free $\uparrow$ & Success $\uparrow$ & Balance $\uparrow$ \\
\midrule
Flow matching          & \studentsecond{0.849}{0.004} & \studentsecond{0.846}{0.038} & \studentsecond{0.949}{0.024} \\
IMLE                   & \studentscore{0.784}{0.011} & \studentscore{0.558}{0.072} & \studentscore{0.621}{0.104} \\
Drifting               & \studentscore{0.364}{0.078} & \studentscore{0.685}{0.071} & \studentscore{0.328}{0.417} \\
\textbf{MoSDOT (ours)} & \studentbest{0.971}{0.003} & \studentbest{0.971}{0.004} & \studentbest{0.963}{0.019} \\
\bottomrule
\end{tabular}
\caption{\textbf{One-step student landmark diagnostics.} Mean $\pm$ one standard deviation over 4 seeds;
same $8192$ noises $\times\,3$ agents per seed as the teacher table. Best
and second-best in \colorbox{tblBest}{\strut sage} and
\colorbox{tblSecond}{\strut powder blue}. Higher is better for all metrics.}
\label{tab:landmark_endpoint_student}
\end{table}
\end{minipage}\par

\par\noindent\begin{minipage}{\linewidth}
\subsection{Full Route-Locality Diagnostics}
\label{app:full_route_locality}

Table~\ref{tab:full_route_locality} reports the full route-locality
diagnostics, combining sample-level and MLP-level measurements. The
sample-level columns (own-cell entropy, mix rate, KNN flip, off-diagonal
AUC excess) are computed on replayed teacher samples and capture whether
nearby source points within an agent's noise slice receive consistent
mode labels. The MLP-level columns (endpoint Jacobian ratio, counterfactual
flip rate) are computed by re-evaluating the restored teacher MLP under
controlled noise perturbations, providing \emph{causal} evidence that the
model itself uses $z_{-i}$ to choose agent~$i$'s branch rather than this
dependence being an artifact of training-sample correlations.

Across both families, MoSDOT is closest to the lowest-entanglement
reference (independent flow matching) and substantially below IMLE and
Drifting, while simultaneously producing the cleanest endpoints
(Table~\ref{tab:landmark_endpoint}). This is the empirical
counterpart of Appendix~\ref{app:fm_block_locality}. When the teacher
objective explicitly couples joint endpoints (IMLE, Drifting),
block-locality is broken and cross-noise dependence appears. When the
coupling is performed upstream by SDOT, endpoint quality improves
\emph{without} reintroducing cross-noise dependence.

\begin{table}[H]
\centering
\footnotesize
\captionsetup{position=bottom,skip=6pt}
\setlength{\tabcolsep}{4pt}
\renewcommand{\arraystretch}{1.2}
\begin{tabular}{@{}lcccccc@{}}
\toprule
\multirow{2}{*}{Method}
  & \multicolumn{4}{c}{\textit{Sample-level}}
  & \multicolumn{2}{c}{\textit{MLP-level}} \\
\cmidrule(lr){2-5}\cmidrule(lr){6-7}
 & $H(m_i\!\mid\!z_i\,\text{cell})$ & Mix rate & KNN flip & AUC excess
 & Jac.\ ratio & CF flip \\
 & \multicolumn{4}{c}{\scriptsize(lower is better)}
 & \multicolumn{2}{c}{\scriptsize(lower is better)} \\
\midrule
Flow  & 0.019\,{\scriptsize$\pm$0.024} & 0.006\,{\scriptsize$\pm$0.010}
      & 0.001\,{\scriptsize$\pm$0.000} & 0.020\,{\scriptsize$\pm$0.008}
      & 0.066\,{\scriptsize$\pm$0.005} & 0.000\,{\scriptsize$\pm$0.000} \\
IMLE  & 0.319\,{\scriptsize$\pm$0.109} & 0.098\,{\scriptsize$\pm$0.036}
      & 0.135\,{\scriptsize$\pm$0.051} & 0.105\,{\scriptsize$\pm$0.042}
      & 0.292\,{\scriptsize$\pm$0.055} & 0.129\,{\scriptsize$\pm$0.050} \\
Drift & 0.666\,{\scriptsize$\pm$0.252} & 0.244\,{\scriptsize$\pm$0.117}
      & 0.315\,{\scriptsize$\pm$0.134} & 0.290\,{\scriptsize$\pm$0.123}
      & 0.772\,{\scriptsize$\pm$0.148} & 0.307\,{\scriptsize$\pm$0.128} \\
\textbf{Ours} & 0.064\,{\scriptsize$\pm$0.024} & 0.020\,{\scriptsize$\pm$0.010}
      & 0.011\,{\scriptsize$\pm$0.003} & 0.012\,{\scriptsize$\pm$0.003}
      & 0.046\,{\scriptsize$\pm$0.015} & 0.002\,{\scriptsize$\pm$0.002} \\
\bottomrule
\end{tabular}
\caption{Full route-locality diagnostics. Sample-level metrics use
12~points per method (4~seeds~$\times$~3~agents). $H$ denotes own-noise
cell entropy with 10~bins, Mix denotes own-cell mix rate, KNN flip uses
$k{=}20$, and AUC excess reports the maximum off-diagonal LDA AUC minus
$0.5$. MLP-level metrics report 4~points for the Jacobian ratio
(4~seeds, agent-averaged) and 12~points for CF flip (4~seeds~$\times$~3~agents).
All values are mean\,$\pm$\,std.}
\label{tab:full_route_locality}
\end{table}
\end{minipage}\par

\noindent\begin{minipage}{\linewidth}
\subsection{Metric Sensitivity Analysis}
\label{app:metric_sensitivity}

Table~\ref{tab:metric_sensitivity_perseed} reports own-cell entropy and
KNN flip across bin counts $b\in\{5,10,20\}$ and neighborhood sizes
$k\in\{10,20,50\}$ for the four diagnostic comparators. The relative
ordering Flow~$\le$~Ours~$\ll$~IMLE~$<$~Drift is preserved across all
hyperparameter settings, ruling out the hypothesis that the
route-locality result depends on a single grid choice.

\begin{table}[H]
\centering
\footnotesize
\captionsetup{position=bottom,skip=6pt}
\setlength{\tabcolsep}{4pt}
\renewcommand{\arraystretch}{1.15}
\begin{tabularx}{\linewidth}{@{}l*{6}{>{\centering\arraybackslash}X}@{}}
\toprule
Method & $H,\,b{=}5$ & $H,\,b{=}10$ & $H,\,b{=}20$
       & KNN $k{=}10$ & KNN $k{=}20$ & KNN $k{=}50$ \\
\midrule
Flow & 0.109\,{\scriptsize$\pm$0.061} & 0.021\,{\scriptsize$\pm$0.024}
     & 0.005\,{\scriptsize$\pm$0.006} & 0.000\,{\scriptsize$\pm$0.000}
     & 0.001\,{\scriptsize$\pm$0.000} & 0.001\,{\scriptsize$\pm$0.001} \\
IMLE & 0.360\,{\scriptsize$\pm$0.101} & 0.319\,{\scriptsize$\pm$0.109}
     & 0.291\,{\scriptsize$\pm$0.106} & 0.134\,{\scriptsize$\pm$0.051}
     & 0.135\,{\scriptsize$\pm$0.051} & 0.137\,{\scriptsize$\pm$0.051} \\
Drift & 0.690\,{\scriptsize$\pm$0.240} & 0.666\,{\scriptsize$\pm$0.252}
      & 0.634\,{\scriptsize$\pm$0.252} & 0.314\,{\scriptsize$\pm$0.134}
      & 0.315\,{\scriptsize$\pm$0.134} & 0.316\,{\scriptsize$\pm$0.133} \\
\textbf{Ours} & 0.181\,{\scriptsize$\pm$0.015} & 0.064\,{\scriptsize$\pm$0.024}
     & 0.032\,{\scriptsize$\pm$0.010} & 0.008\,{\scriptsize$\pm$0.003}
     & 0.011\,{\scriptsize$\pm$0.003} & 0.019\,{\scriptsize$\pm$0.003} \\
\bottomrule
\end{tabularx}
\caption{Route-locality sensitivity. Each entry reports
12~points per method (4~seeds~$\times$~3~agents), mean\,$\pm$\,std. The
relative ranking is preserved across all bin counts and neighborhood sizes.}
\label{tab:metric_sensitivity_perseed}
\end{table}
\end{minipage}\par

\end{document}